\documentclass[conference]{IEEEtran}
\usepackage{amssymb}
\usepackage{cite}
  \usepackage[pdftex]{graphicx}
\usepackage{epstopdf}
\graphicspath{{./figures/}}

\usepackage[cmex10]{amsmath}
\usepackage{algorithmic}
\usepackage{array}
\usepackage{dsfont}
\usepackage{enumerate}
\usepackage{enumitem}
\usepackage{bbm}
\usepackage{subfigure}   

\newtheorem{theorem}{Theorem}[section]

\newcommand{\qed}{\nobreak \ifvmode \relax \else
      \ifdim\lastskip<1.5em \hskip-\lastskip
      \hskip1.5em plus0em minus0.5em \fi \nobreak
      \vrule height0.75em width0.5em depth0.25em\fi}

\def\bx{{\mathbf x}}

\def\bX{{\mathbf X}}
      
\def\Reals{{\mathbb R}}
\def\btheta{{\boldsymbol \theta}}
\def\bmu{{\boldsymbol \mu}}
\def\bomega{{\boldsymbol \omega}}
\def\bgamma{{\boldsymbol \gamma}}

\def\bq{{\mathbf{q}}}

\def\bP{{\mathbf{P}}}
\def\bI{{\mathbf{I}}}

\def\bbM{{\mathbb M}}
\def\bt{{\mathbf{t}}}

\begin{document}
%
\title{Statistical Estimation and Clustering of Group-invariant Orientation Parameters}

\author{
\IEEEauthorblockN{\footnote{Copyright (c) 2012 IEEE. Personal use of this material is permitted. However, permission to use this material for any other purposes must be obtained from the IEEE by sending a request to pubs-permissions@ieee.org.}Yu-Hui Chen\IEEEauthorrefmark{1},
Dennis Wei\IEEEauthorrefmark{2},
Gregory Newstadt\IEEEauthorrefmark{3}, Marc DeGraef\IEEEauthorrefmark{4}, 
Jeffrey Simmons\IEEEauthorrefmark{5} and
Alfred Hero\IEEEauthorrefmark{1}}
\IEEEauthorblockA{\IEEEauthorrefmark{1}University of Michigan, Ann Arbor, MI USA}
\IEEEauthorblockA{\IEEEauthorrefmark{2}IBM Research, Thomas J. Watson Research Center, Yorktown Heights, NY USA}
\IEEEauthorblockA{\IEEEauthorrefmark{3}Google Inc., Pittsburgh, PA USA}
\IEEEauthorblockA{\IEEEauthorrefmark{4}Carnegie Mellon University Pittsburgh, PA USA}
\IEEEauthorblockA{\IEEEauthorrefmark{5}US Air Force Research Laboratory (AFRL), Dayton, OH USA}
}


%


\maketitle

\begin{abstract}
We treat the problem of estimation of orientation parameters whose values are invariant to transformations from a spherical symmetry group. Previous work has shown that any such group-invariant distribution must satisfy a restricted finite mixture representation, which allows the orientation parameter to be estimated using an Expectation Maximization (EM) maximum likelihood (ML) estimation algorithm. In this paper, we introduce two parametric models for this spherical symmetry group estimation problem: 1) the hyperbolic Von Mises Fisher (VMF) mixture distribution and 2) the Watson mixture distribution. We also introduce a new EM-ML algorithm for clustering samples that come from mixtures of group-invariant distributions with different parameters. We apply the models to the problem of mean crystal orientation estimation under the spherically symmetric group associated with the crystal form, e.g., cubic or octahedral or hexahedral. Simulations and experiments establish the advantages of the extended EM-VMF and EM-Watson estimators for data acquired by Electron Backscatter Diffraction (EBSD) microscopy of a polycrystalline Nickel alloy sample. 
\end{abstract}

%
\IEEEpeerreviewmaketitle

\section{Introduction}
\label{sec:intro}
This paper considers estimation of parameters of distributions whose domain is a particular non-Euclidean geometry: a topological space divided into $M$ equivalence classes by actions of a finite spherical symmetry group. A well known example of a finite spherical symmetry group is the point group in 3 dimensions describing the soccer ball, or football, with truncated icosahedral symmetry that also corresponds to the symmetry of the Carbon-60 molecule. This paper formulates a general approach to parameter estimation in distributions defined over such domains. We use a restricted finite mixture representation introduced in~\cite{chen_parameter_2015} for probability distributions that are invariant to actions of any topological group. This representation has the property that the number of mixture components is equal to the order of the group, the distributions in the mixture are all parameterized by the same parameters,  and the mixture coefficients are all equal. This is practically significant since many reliable algorithms have been developed for parameter estimation when samples come from finite mixture distributions~\cite{dempster_maximum_1977,sohn_efficient_2011}.

We apply the representation to an important problem in materials science: analysis of mean orientation in polycrystals. 
Crystal orientation characterizes properties of materials including electrical conductivity and thermal conductivity. Polycrystalline materials are composed of grains of varying size and orientation, where each grain contains crystal forms with similar orientations. The quality of the material is mainly determined by the grain structure i.e. the arrangement of the grains, their orientations, as well as the distribution of the precipitates. Thus accurate estimation of crystal orientation of the grains is useful for predicting how materials fail and what modes of failure are more likely to occur \cite{de_graef_structure_2007}.

The mean orientation of the grain, characterized for example by its Euler angles, can only be specified modulo a set of angular rotations determined by the symmetry group associated with the specific type of crystal, e.g. hexagonal, cubic. This multiplicity of equivalent Euler angles complicates the development of reliable mean orientation estimators. The problem becomes even harder when the orientations are sampled from a region encompassing more than one grain such that the orientations cluster over different mean directions. In such a case, we would like to identify whether the orientations are multi-modally distributed and also estimate the mean direction for each cluster. 

In our previous work~\cite{chen_parameter_2015}, we introduced the finite mixture of Von Mises-Fisher (VMF) distribution for observations that are invariant to actions of a spherical symmetry group. We applied the expectation maximization (EM) maximum likelihood (ML) algorithm, called EM-VMF, to estimate the group-invariant parameters of this distribution. In this paper, we develop a hyperbolic representation simplification of the EM-VMF algorithm that reduces the computation time by a factor of $2$. We also introduce a new group invariant distribution for spherical symmetry groups, called the $\mathcal{G}$-invariant Watson distribution, which like VMF is a density parameterized by location (angle mean) and scale (angle concentration) over the $p$-dimensional sphere. An EM algorithm is presented for estimation of the parameters, called the EM-Watson algorithm. Furthermore, mixture-of-$\mathcal{G}$-invariant Watson (mGIW) and von Mises-Fisher (mGIV) distributions are introduced to perform clustering on the $\mathcal{G}$-invariant sphere. An EM algorithm is presented for estimation of the parameters of the mGIW and mGIV distributions. We illustrate how the Generalized Likelihood Ratio Test (GLRT) can be used to detect the presence of multiple modes in a sample and how it can be combined with the EM algorithm for mGIW and mGIV distributions to cluster multiple orientations on the sphere. 

The performance of the proposed EM orientation estimators is evaluated by simulation and compared to other estimators. The EM orientation estimators are then illustrated on Electron Backscatter Diffraction EBSD data collected from a Nickel alloy whose crystal form induces the $m\overline{3}m$~\cite{newnham_properties_2004} cubic point symmetry group. We establish that the EM orientation estimators result in significantly improved estimates of the mean direction in addition to providing an accurate estimate of concentration about the mean. Furthermore, with the extended mixture models, we are able to identify and cluster multi-modally distributed samples more accurately than the K-means algorithm.

The paper is organized as follows. Section~\ref{sec:group-invariant} describes group invariant random variables and gives the mixture representation for their densities. Section \ref{sec:spherical_symmetry_group} specializes to random variables invariant relative to actions of the spherical symmetry group and develops the $\mathcal G$-invariant VMF and Watson distributions along with  EM-ML parameter estimator. The clustering methods based on the $\mathcal{G}$-invariant distributions along with the GLRT are elaborated in Section \ref{sec:clustering_spherical_symmetry_group}. The crystallography application and data simulation are presented in Section~\ref{sec:app_crystal_orientation_estimation} and the experiment results are shown in Section \ref{sec:experiment}. Section~\ref{sec:conclusion} has concluding remarks.

\section{Group-invariant random variables}
\label{sec:group-invariant}
\def\bfx{{\mathbf x}}
Consider a finite topological group $\mathcal G=\{G_1, \ldots, G_M\}$ of $M$ distinct actions on a topological space $\mathcal X$, $G_i: \mathcal X \rightarrow \mathcal X$ and a binary operation "*" defining the action composition $G_i * G_j$, denoted $G_i G_j$. $\mathcal G$ has the properties that composition of multiple actions is associative, for every action there exists an inverse action, and there exists an identity action \cite{birkhoff_brief_1963}. A real valued function $f(\bx)$ on $\mathcal X$ is said to be invariant under $\mathcal G$ if: $f(G\bx)=f(\bx)$ for $G\in \mathcal G$.   Let $\bX$ be a random variable defined on $\mathcal X$. We have the following theorem for the probability density $f(\bx)$ of $\bX$.
\begin{theorem}
\label{thm:1}
The density function $f: \mathcal X\rightarrow \Reals$ is invariant under $\mathcal G$ if and only if
\begin{eqnarray}
\label{eq:thm1_representation}
\!\begin{aligned}
\exists\ &h: \mathcal X\rightarrow \Reals\ s.t. \\
&f(\bx)= \frac{1}{M} \sum_{i=1}^M h(G_i\bx).
\end{aligned}
\end{eqnarray}
\end{theorem}
This theorem is a slight generalization of \cite[Thm. 2.1]{chen_parameter_2015} in that the density $h(.)$ is not necessarily the same as $f(.)$.  The proof is analogous to that of \cite[Thm. 2.1]{chen_parameter_2015}.

Theorem \ref{thm:1} says that any density $f(\bx)$ that is invariant under group $\mathcal G$ can be represented as a finite mixture of a function and its translates $h(G_i\bx)$ under the group's actions $G_i \in \mathcal G$. As pointed out in~\cite{chen_parameter_2015}, Thm.~\ref{thm:1} has important implications on $\mathcal G$-invariant density estimation and parameter estimation. In particular it can be used to construct maximum likelihood estimators for parametric densities. Let $h(\bx;\btheta)$ be a density on $\mathcal X$ that is parameterized by a parameter $\btheta$ in a parameter space $\Theta$. We extend  $h(\bx;\btheta)$ to a $\mathcal G$-invariant density $f$ by using Thm. \ref{thm:1}, obtaining:
\begin{eqnarray}
f(\bx;\btheta)=\frac{1}{M} \sum_{i=1}^M h_i(\bx;\btheta),
\label{eq:SSG}
\end{eqnarray}
where $h_i(\bx;\btheta)=h(G_i\bx;\btheta)$. This density is of the form of a finite mixture of densities $h_i(\bx;\btheta)$ of known parametric form where the mixture coefficients are all identical and equal to $1/M$. Maximum likelihood (ML) estimation of the parameter $\btheta$ from an i.i.d. sample $\{\bx_i\}_{i=1}^n$ from any $\mathcal G$-invariant density $f$ can now be performed using finite mixture model methods \cite{mclachlan_finite_2004} such as the  Expectation-Maximization (EM) algorithm~\cite{dempster_maximum_1977} or the restricted Boltzman machine (RBM) \cite{sohn_efficient_2011}.

\section{ML within a Spherical Symmetry Group}
\label{sec:spherical_symmetry_group}
As in~\cite{chen_parameter_2015} we specialize Thm.~\ref{thm:1} to estimation of parameters for the case that the probability density is on a sphere and is invariant to actions in a spherical symmetry group. In Section~\ref{sec:app_crystal_orientation_estimation} this will be applied to a crystallography example under spherical distribution likelihood models for the mean crystal orientation. In general, the measured and mean orientations can be represented by Euler angles~\cite{eberly_euler_2008}, Rodrigues Vectors~\cite{rodrigues_lois_1840}, or Quaternions~\cite{altmann_rotations_2005}. As in~\cite{chen_parameter_2015}, we use the quaternion representation to enable orientations to be modeled by spherical distributions since the quaternion representation is a $4$D vector on the $3$-sphere $S^3$, i.e. $\bq = (q_1, q_2, q_3, q_4)$ such that $\|\bq\| = 1$.

Any of the aforementioned orientation representations have inherent ambiguity due to crystal symmetries. For example, if the crystal has cubic symmetry, its orientation is only uniquely defined up to a 24-fold set of proper rotations of the cube about its symmetry axes.
These actions form a point symmetry group, called $432$, a sub-group of $m\overline{3}m$. In quaternion space, since each orientation corresponds to two quaternions with different sign $\{\bq,-\bq\}$, these rotations reflections, and inversions can be represented as a spherical symmetry group $\mathcal{G}$ of quaternionic matrices $\{\bP_1,\ldots,\bP_{M}\}$, with sign symmetry such that $\bP_i=-\bP_{i-M/2}\ \forall M/2<i\le M$, where $M=48$ for cubic symmetry. 

Based on the symmetry group $\mathcal{G}$, we can define the distance between two quaternions under $\mathcal{G}$ as:
\begin{equation}
\label{eq:sym_dist}
d_{\mathcal{G}}(\bq_1,\bq_2) = \min_{\bP\in\mathcal{G}} \arccos{(\bq_1^T\bP\bq_2)}
\end{equation}
Two quaternions $\bq_1,\bq_2$ are called symmetry-equivalent to each other if they are mapped to an equivalent orientation under $\mathcal{G}$, i.e. $d_{\mathcal{G}}(\bq_1,\bq_2)=0$. A fundamental zone (FZ), also called the fundamental domain, is a conic solid subset of the sphere that can be specified to disambiguate any particular orientation $\bx$. However, as will be seen in Sec. \ref{sec:app_crystal_orientation_estimation}, reduction of the entire data sample $\{\bx_i\}_{i=1}^n$ to a FZ destroys information necessary for maximum likelihood estimation: the entire $\mathcal{G}$-invariant density (\ref{eq:SSG}) must be used. In the following two subsections, we introduce two $\mathcal{G}$-invariant spherical distributions: von Mises-Fisher and Watson distributions~\cite{mardia_directional_1999} to model orientations in quaternion space.

\subsection{Hyperbolic $\mathcal G$-invariant von Mises-Fisher Distribution}
\label{sec:ginv_VMF_dist}
First we review the $\mathcal G$-invariant von Mises-Fisher distribution presented in~\cite{chen_parameter_2015}. The von Mises-Fisher (VMF) distribution arises in directional statistics~\cite{mardia_directional_1999} as an analogue of the multivariate Gaussian distribution on the $(p-1)$-dimensional sphere $S^{(p-1)}\subset \Reals^p$, where $p\geq 2$. The VMF distribution is parameterized by the mean direction $\bmu\in S^{(p-1)}$ and the concentration parameter $\kappa\ge 0$:
\begin{equation}
\phi(\mathbf{x};\mathbf{\bmu},\kappa) = c_p(\kappa)\exp{(\kappa\bmu^T\mathbf{x})},
\label{eq:VMF_pdf}
\end{equation}
where $c_p(\kappa) = \frac{\kappa^{p/2-1}}{(2\pi)^{p/2}I_{p/2-1}(\kappa)}$ and $I_p(\cdot)$ is the modified Bessel function of the first kind of order $p$. Given an i.i.d. sample $\{\bx_i\}_{i=1}^n$ from the VMF distribution, the ML estimator has the closed-form expressions \cite{mardia_directional_1999}
\begin{align}
\label{eq:ML_estimator}
\hat{\bmu}=\frac{\mathbf{\bgamma}}{\|\mathbf{\bgamma}\|},\ \hat{\kappa}=A_p^{-1}\left(\frac{\|\mathbf{\bgamma}\|}{n}\right),\
\end{align}
where $\mathbf{\bgamma}=\sum_{i=1}^{n}\mathbf{\bx}_i$ and $A_p(u)=\frac{I_{p/2}(u)}{I_{p/2-1}(u)}$.

Let $\mathcal{G}$ be a group of symmetric actions $\{\bP_1, \ldots, \bP_M\}$ acting on the quaternionic representation of orientation on the $3$-dimensional sphere $S^{3}$. We extend the VMF distribution (\ref{eq:VMF_pdf}) using the mixture representation in Thm~\ref{thm:1}:
\begin{eqnarray}
\label{eq:mixture_densityp}
f_v(\mathbf{x};\bmu,\kappa)
&=&\frac{1}{M}\sum_{m=1}^M\phi(\bP_m\mathbf{x};\bmu, \kappa)\\
\label{eq:mixture_density}
&=&\frac{1}{M}\sum_{m=1}^M\phi(\mathbf{x};\bP_m\bmu, \kappa)
\end{eqnarray}
where in going from (\ref{eq:mixture_densityp}) to (\ref{eq:mixture_density}) we used the inner product form $\bmu^T \bx$ in (\ref{eq:VMF_pdf}) and the symmetry of $\bP_m$. The expression (\ref{eq:mixture_density}) for the extended VMF distribution is in the form of
a finite mixture of standard VMF distributions on the same random variable $\bx$ having different mean parameters $\bmu_m =\bP_m\bmu$ but having the same concentration parameter $\kappa$.

The finite mixture (\ref{eq:mixture_density}) for the $\mathcal G$-invariant density $f_v(\mathbf{x};\bmu,\kappa)$ is in a form for which an EM algorithm~\cite{dempster_maximum_1977} can be implemented to compute the ML estimates of $\bmu$ and $\kappa$. Denoting the parameter pair as $\bomega=\{\bmu,\kappa\}$, the EM algorithm generates a sequence $\{\bomega^{(k)}\}$ of estimates that monotonically increase the likelihood. These estimates are given by $\bomega^{(k+1)}= \arg\max_{\bomega}  E_{S|\bX,\bomega^{(k)}}[\log{L(\bomega;\{\bx_i,s_i\})}]$, where $s_i$ is a latent variable assigning $\bx_i$ to a particular mixture component in (\ref{eq:mixture_density}) and $L(\bomega;\{\bx_i, s_i\})$ is the likelihood function of $\bomega$ given the complete data $\{\bx_i, s_i\}_{i=1}^n$. Specifically,
\begin{eqnarray}
\label{eq:qfunction}
&&E_{S|\bX,\bomega}[\log{L(\bomega;\{\bx_i,s_i\})}] \\
&=& \sum_{i=1}^{n}\sum_{m=1}^Mr_{i,m}(\log{c_p(\kappa)}+\kappa(\bP_m\bmu)^T\mathbf{x}_i),
\nonumber
\end{eqnarray}
where $r_{i,m}=P(s_i=m|\bx_i,\bomega)$. The EM algorithm takes the form:

E-step:
\begin{equation}
\label{eq:VMF_Estep}
\begin{split}
r_{i,m} = \frac{\phi(\bx_i; \bP_m\bmu, \kappa)}{\sum_{l=1}^M\phi(\bx_i; \bP_l\bmu, \kappa)}, m\in\left\{1,2,\ldots,M\right\}.
\end{split}
\end{equation}

M-step:
\begin{align}
\label{eq:EM_Mstep}
\hat{\bmu}&=\frac{\bgamma}{\|\bgamma\|},\ \hat{\kappa}=A_p^{-1}\left(\frac{\|\bgamma\|}{n}\right),  \\
\bgamma&=\sum_{i=1}^{n}\sum_{m=1}^Mr_{i,m}\bP_m^T\bx_i. \label{eq:VMF_Mstep_gamma}
\end{align}

The hyberbolic $\mathcal G$-invariant von Mises-Fisher distribution is obtained by exploiting the sign symmetry in $\mathcal G$. In particular, (\ref{eq:VMF_Mstep_gamma}) in the M-step can be re-written as:
\begin{equation}
\label{eq:EM_Mstep_simplified}
\begin{split}
\bgamma&=\sum_{i=1}^{n}\sum_{m=1}^Mr_{i,m}\bP_m^T\mathbf{x}_i \\
&=\sum_{i=1}^{n}\left(\sum_{m=1}^{M/2}r_{i,m}\bP_m^T\bx_i - r_{i,\frac{M}{2}+m}\bP_m^T\bx_i\right) \\
&=\sum_{i=1}^{n}\sum_{m=1}^{M/2} \frac{\sinh{(\kappa(\bP_m\bmu)^T\bx_i)}}{\sum_{l=1}^{M/2}\cosh{(\kappa(\bP_l\mu)^T\bx_i)} } \bP_m^T\bx_i,
\end{split}
\end{equation}
where $\sinh,\cosh$ are the hyperbolic sinusoidal functions. Equation (\ref{eq:VMF_Estep}) in E-step is simplified as:
\begin{equation}
\label{eq:EM_Estep_simplified}
r'_{i,m} = \frac{\sinh{(\kappa(\bP_m\bmu)^T\bx_i)}}{\sum_{l=1}^{M/2}\cosh{(\kappa(\bP_l\mu)^T\bx_i)}}, m\in\left\{1,2,\ldots,\frac{M}{2}\right\}.
\end{equation}

In Section~\ref{sec:experiment} we demonstrate the computational improvement of the hyperbolic form of the EM algorithm obtained by substituting (\ref{eq:EM_Mstep_simplified}), (\ref{eq:EM_Estep_simplified}) into (\ref{eq:EM_Mstep}), (\ref{eq:VMF_Estep}) respectively.

\subsection{$\mathcal G$-invariant Watson Distribution}
\label{sec:ginv_Watson_dist}
As described at the beginning of this section, each orientation corresponds to two quaternions with different sign, which is equivalent to an axis of the sphere. For axial data it is more natural to use the Watson distribution~\cite{watson_equatorial_1965}, which models the probability distribution of axially symmetric vectors on the $(p-1)$-dimensional unit sphere, i.e. $\pm\bx\in S^{p-1}$ are equivalent. Similar to VMF, the distribution is parametrized by a mean direction $\pm\bmu\in S^{p-1}$, and a concentration parameter $\kappa\in\Reals$ that is no longer necessarily non-negative. Its probability density function is
\begin{equation}
\label{eq:Watson_pdf}
W_p(\bx; \bmu,\kappa) = \frac{1}{\bbM(\frac{1}{2},\frac{p}{2},\kappa)}\exp{\left(\kappa(\bmu^T\bx)^2\right)},
\end{equation}
where $\bbM$ is the Kummer confluent hypergeometric function defined in~\cite{bateman_higher_1955}. According to (\ref{eq:Watson_pdf}), the positive-negative pair of group actions $\{\bP_m,-\bP_m\}$ contribute the same value in the density function. The set of the group action pairs $\mathcal{G}'=\{\{\bP_m,-\bP_m\}\}_{m=1}^{M/2}$ is the quotient group $\mathcal{G}/\mathcal{I}$, where $\mathcal{I}=\{\bI_p,-\bI_p\}\subset\mathcal{G}$ and $\bI_p$ is the identity matrix of dimension $p$. Therefore, $\mathcal{G}'$ is also a group and we can use Thm~\ref{thm:1} to extend the Watson distribution to the mixture representation under $\mathcal{G}'$:
\begin{equation}
\label{eq:Watson_mixture_density}
f_w(\bx;\bmu,\kappa) = \frac{1}{M'} \sum_{m=1}^{M'} W_p(\bx; \bP_m\bmu,\kappa),
\end{equation}
where $M'=M/2$. The ML estimates of $\bmu$ and $\kappa$ can also be calculated by the EM algorithm. The E-step for the Watson mixture distribution is
\begin{equation}
\label{eq:Watson_EM_Estep}
\begin{split}
r_{i,m} = \frac{\exp{\{\kappa((\bP_m\bmu)^T\mathbf{x}_i)^2\}}}{\sum_{l=1}^{M'}\exp{\{\kappa((\bP_l\bmu)^T\mathbf{x}_i)^2\}}},m\in\left\{1,2,...,M'\right\}.
\end{split}
\end{equation}

For the M-step, we take a similar approach as~\cite{mardia_directional_1999} as follows:
\begin{equation}
\label{eq:Watson_Mstep_Dev}
\begin{split}
&E_{S|\bX,\bomega}[\log{L(\bomega;\{\bx_i,s_i\})}] \\
=& n\left( \kappa\bmu^T\tilde{T}\bmu - \log{\left(M'\bbM\left(\frac{1}{2},\frac{p}{2},\kappa\right)\right)}\right),
\end{split}
\end{equation}
where $\tilde{T}=\frac{1}{n}\sum_{i=1}^n\sum_{m=1}^{M'}r_{i,m}(\bP_m^T\bx_i\bx_i^T\bP_m)$ is the scatter matrix of $\bx_1,...,\bx_n$. Let $\tilde{t}_1,...,\tilde{t}_p$ be the eigenvalues of $\tilde{T}$ with 
\begin{equation}
\tilde{t}_1\ge...\ge\tilde{t}_p,
\end{equation}
and let $\pm\bt_1,...,\pm\bt_p$ be the corresponding unit eigenvectors.
Since we want to find $\bmu$ which maximizes (\ref{eq:Watson_Mstep_Dev}) such that $\bmu^T\bmu=1$, the estimator of $\bmu$ for fixed $\kappa$ has the following form:
\begin{equation}
\begin{split}
\label{eq:Watson_mu_est}
\hat\bmu &= \bt_1, \hat{\kappa} > 0, \\
\hat\bmu &= \bt_p, \hat{\kappa} < 0.
\end{split}
\end{equation}

Similarly by fixing $\bmu$ and setting to zero the derivative of (\ref{eq:Watson_Mstep_Dev}) with respect to $\kappa$, we have:
\begin{equation}
\label{eq:Watson_Tp_func}
\begin{split}
&Y_p(\kappa) = \frac{\bbM'(\frac{1}{2},\frac{p}{2},\kappa)}{\bbM(\frac{1}{2},\frac{p}{2},\kappa)} = \frac{\sum_{i=1}^n\sum_{m=1}^{M'}r_{i,m}(\bmu^T\bP_m^T\bx_i)^2}{n} \\
\Rightarrow& \hat{\kappa} = Y_p^{-1}\left( \frac{\sum_{i=1}^n\sum_{m=1}^{M'}r_{i,m}(\bmu^T\bP_m^T\bx_i)^2}{n}\right),
\end{split}
\end{equation}
The final estimates of $\bmu$ and $\kappa$ are obtained by checking both cases ($\hat{\kappa}>0$, $\hat{\kappa}<0$) and choosing the one which is consistent for (\ref{eq:Watson_mu_est})(\ref{eq:Watson_Tp_func}).

\section{Clustering with a Spherical Symmetry Group}
\label{sec:clustering_spherical_symmetry_group}
In this section we extend the parameter estimation problem to the situation where there are multiple group-invariant distributions with different parameters that govern the samples. This problem arises, for example, in poly-crystaline materials when estimating the mean crystal orientation over a region containing more than one grain (perhaps undetected). This problem can be solved by first applying some standard clustering methods, e.g. K-means\cite{hartigan_algorithm_1979}, and then estimating the parameters for each cluster. However, clustering methods based on the distance relation between the samples are complicated by the presence of spherical symmetry because it is necessary to distinguish modes that are due only to symmetry from those that distinguish different clusters. Therefore, we propose a model-based clustering algorithm which accommodates symmetry to handle this problem.

Consider the situation where the samples $\{\bx_i\}_{i=1}^n$ follow a mixture of $\mathcal{G}$-invariant density functions. For the VMF distribution, the mixture density has the following form:
\begin{equation}
\label{eq:mixture_of_Ginv_VMF}
g_v(\bx;\{\bmu_c,\kappa_c,\alpha_c\}) = \sum_{c=1}^C\alpha_c\left(\sum_{m=1}^M \frac{1}{M} \phi(\bx;\bP_m\bmu_c,\kappa_c)\right),
\end{equation}
where $C$ is the number of clusters assumed to be fixed a priori, $\bmu_c,\kappa_c$ are the parameters for the $c$-th cluster and $\alpha_c$ are the mixing coefficients where $\sum_{c=1}^C\alpha_c=1$ and $\alpha_c>0$ for all $c$. The parameters of (\ref{eq:mixture_of_Ginv_VMF}) can be estimated by the EM algorithm:

E-step:
\begin{equation}
\label{eq:mVMF_mClusters_Estep}
r_{i,c,m}=\frac{\alpha_c\phi(\bx_i; \bP_m\bmu_c,\kappa_c)}{\sum_{h=1}^C\alpha_{h}\sum_{l=1}^M\phi(\bx_i;\bP_{l}\mu_{h}, \kappa_{h})}
\end{equation} 

M-step:
\begin{align}
\alpha_c&=\sum_{i=1}^n\sum_{m=1}^Mr_{i,c,m}, \hat{\bmu}_c=\frac{\bgamma_c}{\|\bgamma_c\|}, \hat{\kappa}_c=A_p^{-1}\left(\frac{\|\bgamma_c\|}{n\alpha_c}\right),\\
\label{eq:mVMF_mClusters_Mstep}
\bgamma_c&=\sum_{i=1}^n\sum_{m=1}^M r_{i,c,m}\bP_m^T\bx_i,
\end{align}
where $r_{i,c,m}$ is the probability of sample $\bx_i$ belonging to the $c$-th cluster and the $m$-th symmetric component.

For the Watson distribution, the mixture of $\mathcal{G}$-invariant Watson density is
\begin{equation}
\label{eq:mixture_of_Ginv_Watson}
g_w(\bx;\{\bmu_c,\kappa_c, \alpha_c\}) = \sum_{c=1}^C\alpha_c\left(\sum_{m=1}^{M'} \frac{1}{M'}W_p(\bx; \bP_m\bmu_c,\kappa_c)\right)
\end{equation}

The E-step is similar to (\ref{eq:mVMF_mClusters_Estep}) with $\phi$ replaced by $W_p$ function. The M-step can be computed with a similar approach as in Section~\ref{sec:ginv_Watson_dist} with the following modifications:
\begin{align}
\label{eq:Watson_mClusters_Mstep_hatT}
\tilde{T_c}&=\frac{1}{n\alpha_c}\sum_{i=1}^n\sum_{m=1}^{M'} r_{i,c,m}(\bP_m^T\bx_i\bx_i^T\bP_m), \\
\label{eq:Watson_mClusters_Mstep_kappa}
\hat{\kappa}_c&=Y_p^{-1}\left(\frac{\sum_{i=1}^n\sum_{m=1}^{M'}r_{i,c,m}(\bmu_c^T\bP_m^T\bx_i)^2}{n\alpha_c}\right),
\end{align}
where $\alpha_c=\sum_{i=1}^n\sum_{m=1}^Mr_{i,c,m}$.
\subsection{Multi-modality Tests on $\mathcal{G}$-invariant Spherical Distributions}
Given sample set $\{\bx_i\}_{i=1}^n$ on $S^{p-1}$, the objective is to determine whether the $n$ samples are drawn from one single distribution or a mixture of $C$ distributions. For polycrystalline materials, the result of this determination can be used to discover undetected grains within a region. We propose to use a multi-modal hypothesis test based on the $\mathcal{G}$-invariant distributions to solve this problem. The two hypotheses are $H_0$: The samples are from a single $\mathcal{G}$-invariant distribution $f(\bx;\{\bmu,\kappa\})$; and $H_1$: The samples are from a mixture of $C$ distributions $g(\bx;\{\bmu_c,\kappa_c,\alpha_c\}_{c=1}^C)$. The Generalized Likelihood Ratio Test (GLRT)~\cite{hero_statistical_2000} has the following form:
\begin{equation}
\label{eq:multi_sample_GLRT}
\begin{aligned}
\Lambda_{GLR} &= \frac{\max_{\{\bmu_c,\kappa_c,\alpha_c\}_{c=1}^C\in\Theta_1}g(\{\bx_i\}_{i=1}^n;\{\bmu_c,\kappa_c,\alpha_c\}_{c=1}^C)}{\max_{\{\bmu,\kappa\}\in\Theta_0}f(\{\bx_i\}_{i=1}^n;\{\bmu,\kappa\})} \\
&\gtrless^{H_1}_{H_0} \eta
\end{aligned}
\end{equation}
where $\Theta_0,\Theta_1$ are the parameter spaces for the two hypotheses. The $f$ and $g$ functions for VMF and Watson distributions are defined in (\ref{eq:mixture_density}), (\ref{eq:Watson_mixture_density}) and (\ref{eq:mixture_of_Ginv_VMF}), (\ref{eq:mixture_of_Ginv_Watson}) respectively and the test statistic $\Lambda_{GLR}$ can be calculated by the proposed EM algorithm. According to Wilks's theorem~\cite{wilks_large-sample_1938} as $n$ approaches $\infty$, the test statistic $2\log{\Lambda_{GLR}}$ will be asymptotically $\chi^2$-distributed with degrees of freedom equal to $(p+1)(C-1)$, which is the difference in dimensionality of $\Theta_0$ and $\Theta_1$. Therefore, the threshold $\eta$ in (\ref{eq:multi_sample_GLRT}) can be determined by a given significance level $\alpha$.

\section{Application to Crystallographic Orientation}
\label{sec:app_crystal_orientation_estimation}
Crystal orientation and the grain distribution in polycrystalline materials determine the mechanical properties of the material, such as, stiffness, elasticity, and deformability. Locating the grain regions and estimating their orientation and dispersion play an essential role in detecting anomalies and vulnerable parts of materials.

Electron backscatter diffraction (EBSD) microscopy acquires crystal orientation at multiple locations within a grain by capturing the Kikuchi diffraction patterns of the backscatter electrons ~\cite{saruwatari_crystal_2007}. A Kikuchi pattern can be translated to crystal orientation through Hough Transformation analysis~\cite{lassen_automated_1994} or Dictionary-Based indexing~\cite{park_ebsd_2013}. The process of assigning mean orientation values to each grain is known as indexing. Crystal forms  possess  point symmetries, e.g. triclinic, tetragonal, or cubic, leading to  a probability density of measured orientations that is invariant over an associated spherical symmetry group $\mathcal{G}$. Therefore, when the type of material has known symmetries, e.g., cubic-type symmetry for nickel or gold, the $\mathcal{G}$-invariant VMF and Watson models introduced in Section~\ref{sec:spherical_symmetry_group} can be applied to estimate the mean orientation $\bmu_g$ and the concentration $\kappa_g$ associated with each grain. Furthermore, the clustering method along with the multi-sample hypothesis test in Section~\ref{sec:clustering_spherical_symmetry_group} can be used to detect the underlying grains within a region.

\subsection{Simulation of Crystallographic Orientation}
\label{sec:simulation_orientations}
To simulate the crystallographic orientations, we first draw random samples from VMF and Watson distributions with $p=4$. The random variable $\bx$ in a spherical distribution can be decomposed~\cite{mardia_directional_1999}:
\begin{equation}
\label{eq:normal_tangent_decompose}
\bx=t\bmu+\sqrt{1-t^2}S_\bmu(\bx),
\end{equation}
where $t=\bmu^T\bx$ and $S_\bmu(\bx)=(I_p-\bmu\bmu^T)\bx/\|(I_p-\bmu\bmu^T)\bx\|$. Let $f(\bx;\bmu)$ be the p.d.f. of the distribution where $\bmu$ is the mean direction. According to the normal-tangent decomposition property, for any rotationally symmetric distribution, $S_\bmu(\bx)$ is uniformly distributed on $S_{\bmu^\bot}^{p-2}$, the $(p-2)$-dimensional sphere normal to $\bmu$, and the density of $t=\bx^T\bmu$ is given by:
\begin{equation}
\label{eq:tangent_density}
t\mapsto cf(t)(1-t^2)^{(p-3)/2}.
\end{equation}

For VMF distribution, substituting (\ref{eq:normal_tangent_decompose}) into (\ref{eq:VMF_pdf}) and combining with (\ref{eq:tangent_density}), we have the density of the tangent component $t$ as:
\begin{equation}
\label{eq:VMF_tangent_density}
\begin{split}
f_v(t)&= C_v\exp{\{\kappa t\}}(1-t^2)^{(p-3)/2} \\
C_v&=\left(\frac{\kappa}{2}\right)^{(p/2-1)}\left(I_{p/2-1}(\kappa)\Gamma\left(\frac{p-1}{2}\right)\Gamma\left(\frac{1}{2}\right)\right)^{-1}.
\end{split}
\end{equation}

Similarly, the density of the tangent component of Watson distribution is:
\begin{equation}
\label{eq:Watson_tangetn_density}
\begin{split}
f_w(t)&= C_w\exp{\{\kappa t^2\}}(1-t^2)^{(p-3)/2} \\
C_w&=\frac{\Gamma(\frac{p}{2})}{\Gamma(\frac{p-1}{2})\Gamma(\frac{1}{2})}\frac{1}{\bbM(\frac{1}{2},\frac{p}{2},\kappa)}.
\end{split}
\end{equation}
Random samples from the density functions (\ref{eq:VMF_tangent_density}) and (\ref{eq:Watson_tangetn_density}) can be easily generated by rejection sampling. 

The generated quaternions from VMF and Watson distributions are then mapped into the Fundamental Zone (FZ) with the symmetric group actions to simulate the wrap-around problem we observe in real data, i.e. observations are restricted to a single FZ. For cubic symmetry, the FZ in quaternion space is defined in the following set of equations: \\
\begin{minipage}{0.48\linewidth}
\begin{equation}
\begin{cases}
|q_2/q_1|\le\sqrt{2}-1 \\
|q_3/q_1|\le\sqrt{2}-1 \\
|q_4/q_1|\le\sqrt{2}-1 \\
|q_2/q_1 + q_3/q_1 + q_4/q_1|\le 1 \\
|q_2/q_1 - q_3/q_1 + q_4/q_1|\le 1 \\
|q_2/q_1 + q_3/q_1 - q_4/q_1|\le 1 \\
|q_2/q_1 - q_3/q_1 - q_4/q_1|\le 1 \nonumber
\end{cases}
\end{equation}
\end{minipage}
\begin{minipage}{0.48\linewidth}
\begin{equation}
\label{eq:FZ_equations}
\begin{cases}
|q_2/q_1 - q_3/q_1|\le\sqrt{2} \\
|q_2/q_1 + q_3/q_1|\le\sqrt{2} \\
|q_2/q_1 - q_4/q_1|\le\sqrt{2} \\
|q_2/q_1 + q_4/q_1|\le\sqrt{2} \\
|q_3/q_1 - q_4/q_1|\le\sqrt{2} \\
|q_3/q_1 + q_4/q_1|\le\sqrt{2} \\
\end{cases}
\end{equation}
\end{minipage}
where $q_i$ is the $i$-th component of quaternion $\bq$.

\section{Experimental Results}
\label{sec:experiment}
\subsection{$\mathcal G$-invariant EM-ML Parameter Estimation on Simulated Data}
\begin{figure}
  \centering
  \subfigure[VMF Simulated Data]{
    \label{fig:mu_est_VMFdata} 
    \includegraphics[width=4.25cm]{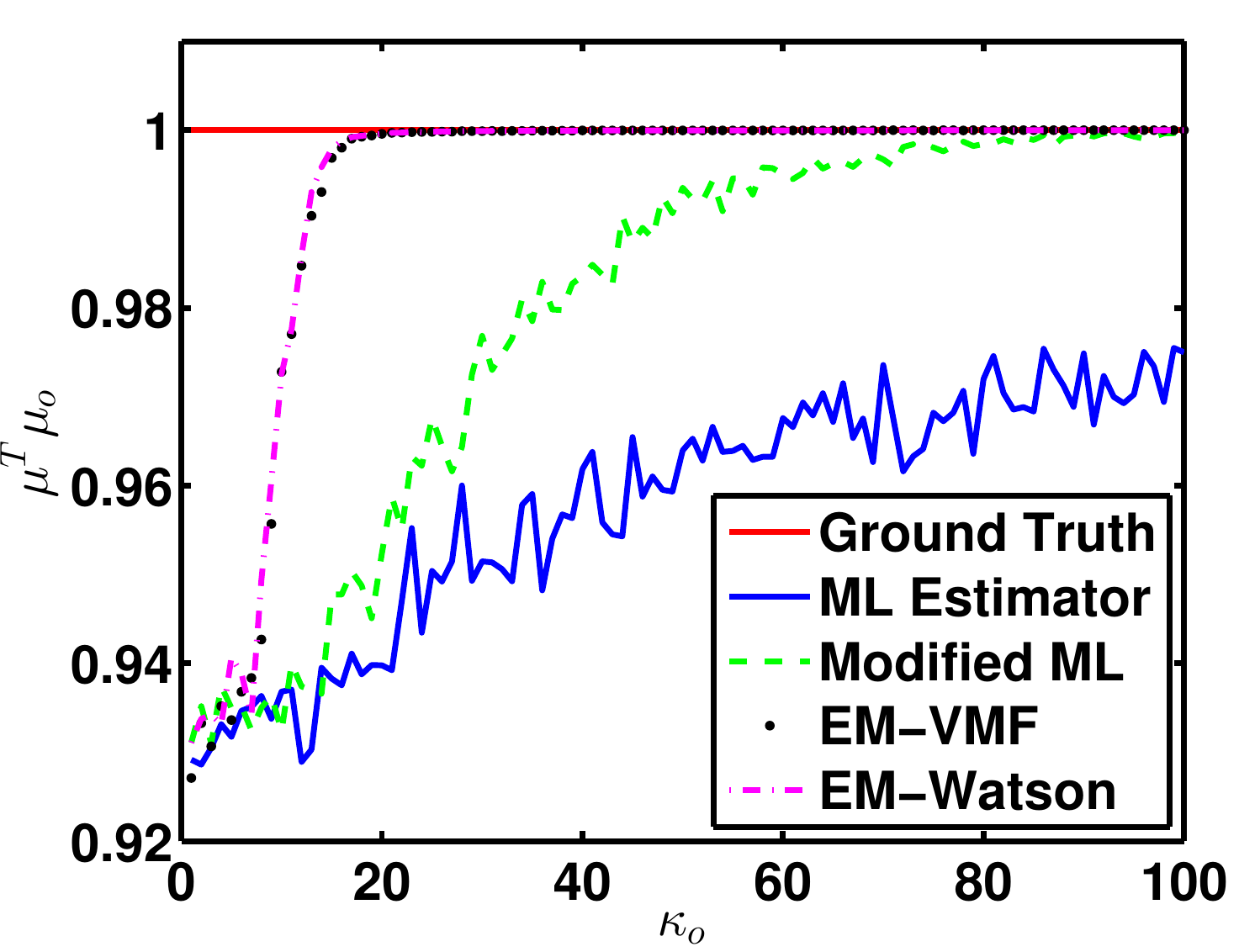}}
  \subfigure[Watson Simulated Data]{
    \label{fig:mu_est_Watsondata} 
    \includegraphics[width=4.25cm]{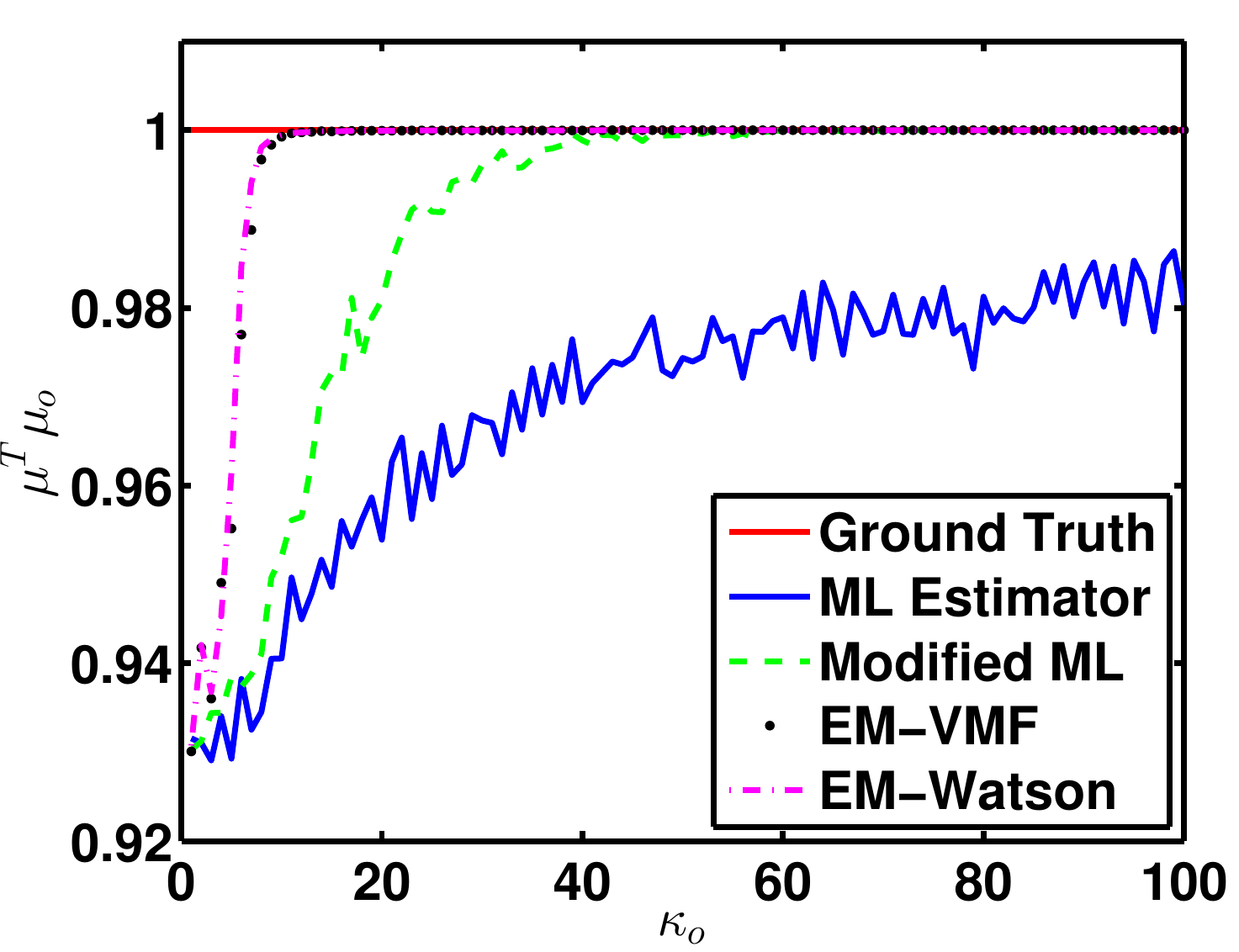}}
  \caption{Mean orientation estimator comparisons for $\mathcal G$-invariant densities. Shown is the average inner product $\bmu_o^T\hat{\bmu}$ of four estimators $\hat{\bmu}$ when $\bmu_o$ is the true mean orientation as a function of the true concentration parameter $\kappa_o$ for the data simulated from VMF (Fig.~\ref{fig:mu_est_VMFdata}) and from Watson (Fig.~\ref{fig:mu_est_Watsondata}) distribution. The naive estimator ("ML Estimator" in blue line) does not attain perfect estimation (inner product $=1$) for any $\kappa_o$ since it does not account for the spherical symmetry group structure. The modified ML (green dashed line) achieves perfect estimation as $\kappa_o$ becomes large.  The proposed EM-ML methods ("EM-VMF", "EM-Watson") achieve perfect estimation much faster than the other methods even under model mismatch (EM-VMF for Watson simulated data and vice versa).}
  \label{fig:mu_est} 
\end{figure}

\begin{figure}
  \centering
  \subfigure[VMF Simulated Data]{
    \label{fig:kappa_est_VMFdata} 
    \includegraphics[width=4.25cm]{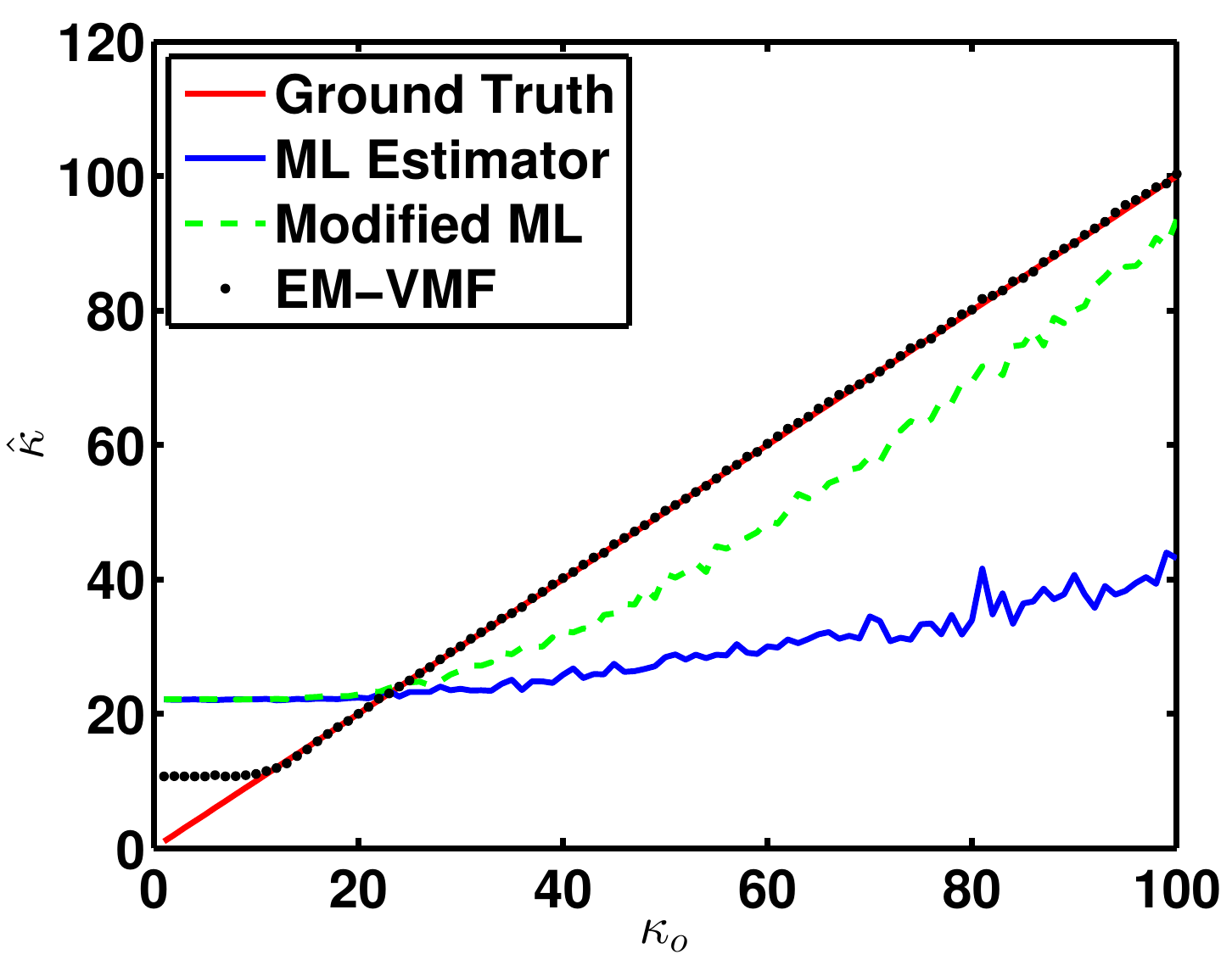}}
  \subfigure[Watson Simulated Data]{
    \label{fig:kappa_est_Watsondata} 
    \includegraphics[width=4.25cm]{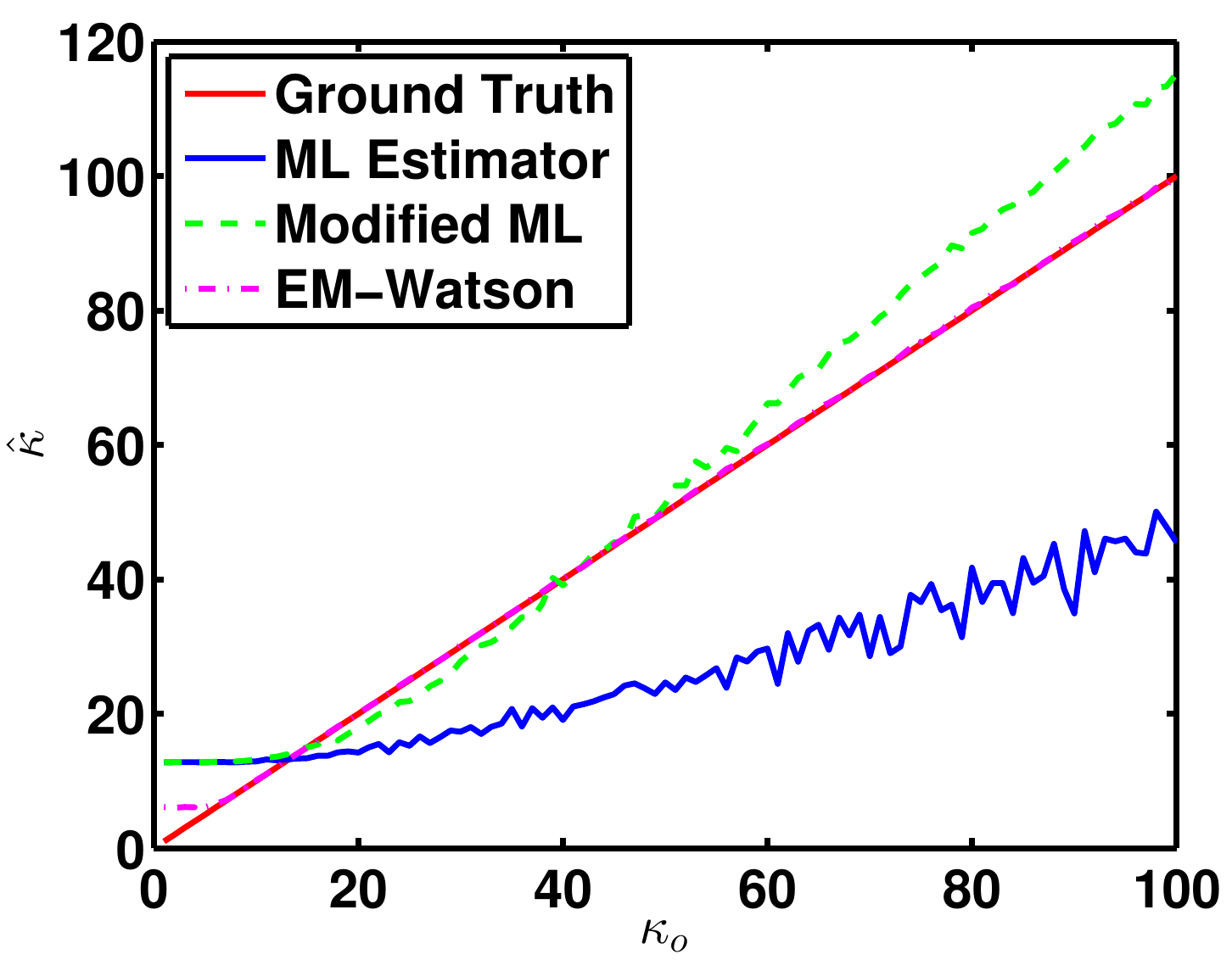}}
  \caption{Concentration parameter estimator bias as a function of the true concentration $\kappa_o$ for data simulated from VMF (Fig.~\ref{fig:kappa_est_VMFdata})\cite{chen_parameter_2015} and from Watson (Fig.~\ref{fig:kappa_est_Watsondata}) distributions. The bias of the naive ML (blue solid line) is large over the full range of $\kappa_o$. The modified ML (green dashed line) estimates $\kappa$ more accurately when $\kappa_o$ is small. Our proposed EM-VMF and EM-Watson estimators (black dotted line and magenta dashed line) have lower bias than the other estimators.}
  \label{fig:kappa_est} 
\end{figure}

Sets of $n$ i.i.d. samples were simulated from the VMF or Watson distributions using the method described in Sec.\ref{sec:simulation_orientations} with given $\bmu=\bmu_o,\kappa=\kappa_o$ for the $m\overline{3}m$ point symmetry group associated with the symmetries of cubic crystal lattice planes. The number of samples for each simulation was set to $n=1000$ and $\kappa_o$ was swept from  $1$ to $100$ while, for each simulation run,  $\bmu_o$ was selected uniformly at random. The experiment was repeated $100$ times and the average values of $\hat{\kappa}$ and the inner product $\hat{\bmu}^T \bmu_o$ are shown in Fig.~\ref{fig:mu_est} and \ref{fig:kappa_est}. In the figures we compare performance for the following methods: (1) the naive ML estimator for the standard VMF or Watson model that does not account for the point group structure (labeled "ML Estimator"). (2) Mapping each of the samples $\bx_i$ toward a reference direction $\bx_{r}$ (randomly selected from $\{\bx_i\}_{i=1}^n$), i.e. $\bx_i\mapsto\bP_m\bx_i$, where $\bP_m=\arg\min_{\bP\in\mathcal{G}} \arccos{(\bx_{r}^T\bP\bx)}$, to remove possible ambiguity. Then performing ML for the standard VMF or Watson distribution (labeled "Modified ML"). (3) Applying our proposed EM algorithm directly to the $n$ samples using the mixture of VMF distribution (\ref{eq:VMF_Estep})-(\ref{eq:VMF_Mstep_gamma}) (labeled "EM-VMF") (4) Applying our proposed EM algorithm to the mixture of Watson distribution (\ref{eq:Watson_EM_Estep})-(\ref{eq:Watson_Tp_func}) (labeled "EM-Watson").


Figure \ref{fig:mu_est} shows the inner product values $\bmu_o^T\hat{\bmu}$. The proposed EM-VMF and EM-Watson estimators have similar performance in that they achieve perfect recovery of the mean orientation ($\bmu_o^T\hat{\bmu}=1$) much faster than the other methods as the concentration parameter $\kappa_o$ increases (lower dispersion of the samples about the mean) no matter whether the data is generated from VMF (Fig.~\ref{fig:mu_est_VMFdata}) or Watson distribution (Fig.~\ref{fig:mu_est_Watsondata}), indicating the robustness of the proposed approaches under model mismatch. Notice that when $\kappa_o$ is small ($\kappa_o<20$ for VMF data and $\kappa_o<10$ for Watson data), none of the methods can accurately estimate the mean orientation. The reason is that when $\kappa_o$ is small the samples become nearly uniformly distributed over the sphere. The threshold $\kappa_o$ value at which performance starts to degrade depends on the choice of point symmetry group and the distribution used to simulate the data. In Fig.~\ref{fig:kappa_est} it is seen that the biases of the proposed EM-VMF~\cite{chen_parameter_2015} and EM-Watson $\kappa$ estimators are significantly lower than that of the other methods compared. While the modified ML performs better than the naive ML estimator, its bias is significantly worse than the proposed EM-VMF and EM-Watson approaches. 

\begin{figure}
  \centering
  \subfigure[VMF Simulated Data]{
    \label{fig:computation_time_VMFdata} 
    \includegraphics[width=4cm]{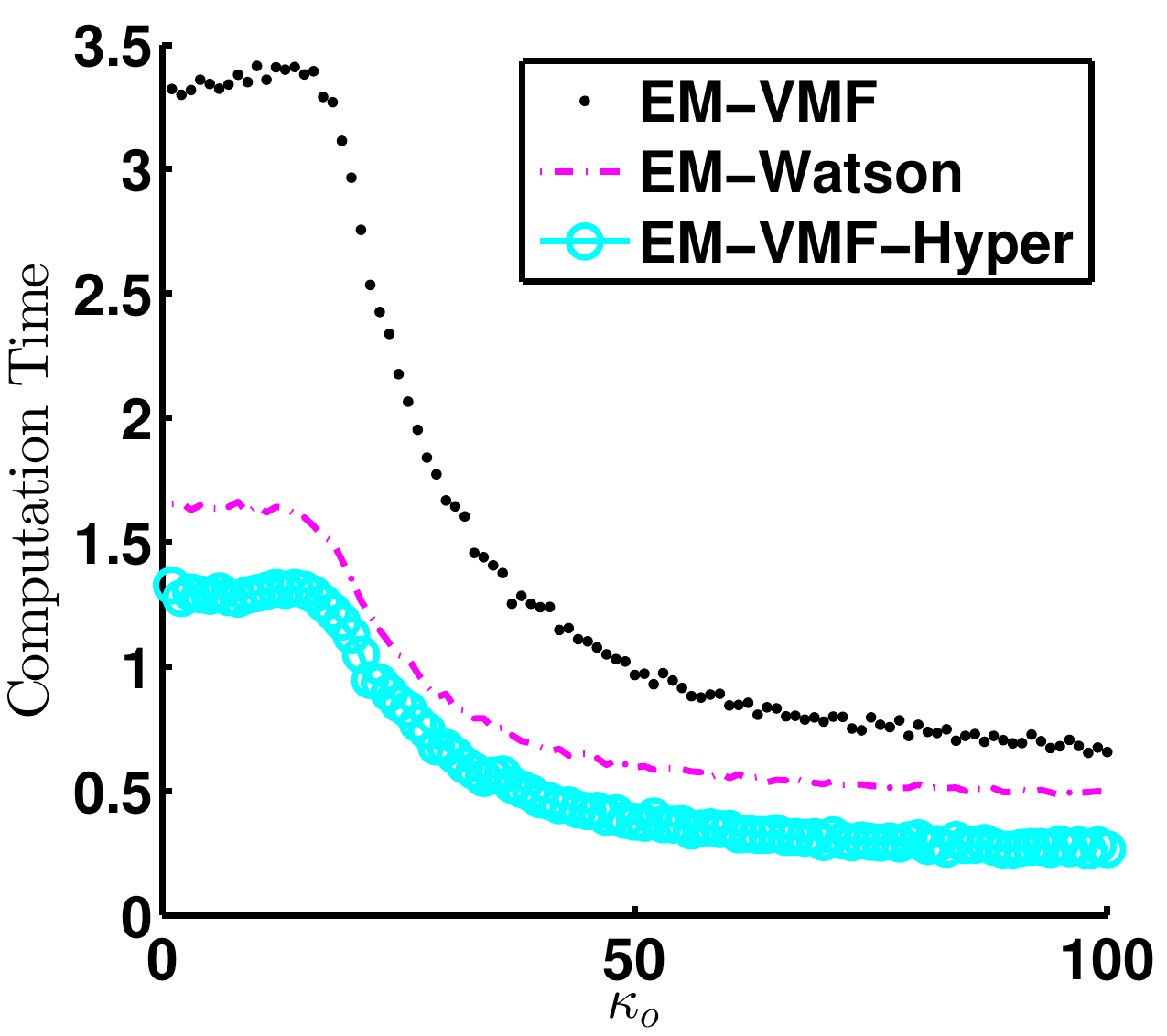}}
  \subfigure[Watson Simulated Data]{
    \label{fig:computation_time_Watsondata} 
    \includegraphics[width=4cm]{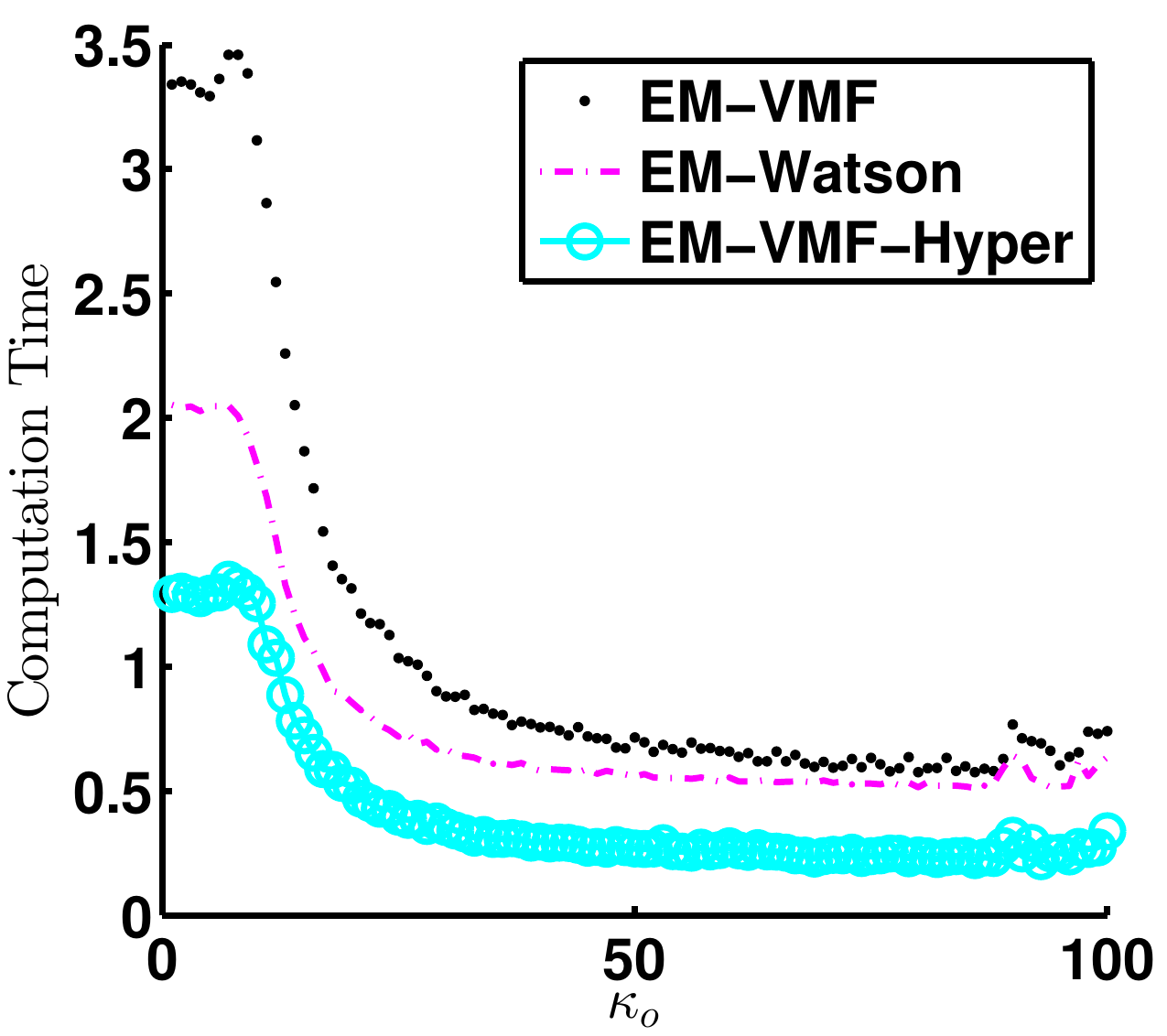}}
  \caption{Computation time for calculating the result in Fig.~\ref{fig:mu_est} and Fig.~\ref{fig:kappa_est}. EM-Watson (magenta dashed line) has less computation time than EM-VMF (black dotted line) because it uses only half of the symmetry operators. EM-VMF-Hyper (cyan circle line) which uses the hyperbolic sinusoidal simplification of EM-VMF reduces the computation time by more than a factor of $2$.}
  \label{fig:computation_time} 
\end{figure}

Figure \ref{fig:computation_time} shows the computation time of the estimation algorithms presented in Fig.~\ref{fig:mu_est} and Fig.~\ref{fig:kappa_est}. The computation time for all methods decreases as $\kappa_o$ becomes larger. When $\kappa_o$ is small ($\kappa_o<20$ for VMF data and $\kappa_o<10$ for Watson data), because the samples are almost uniformly distributed around the sphere, it is difficult for the EM algorithms to converge to the optimal solution and they therefore require maximum number of iterations to stop, forming the plateaus in Fig.~\ref{fig:computation_time}. Notice that EM-Watson requires less time than EM-VMF even though it has more complicated E and M-steps. The reason is that EM-Watson uses only half of the symmetry operators, which corresponds to the size of the quotient group $\mathcal{G}/\mathcal{I}$ as described in Section~\ref{sec:ginv_Watson_dist}. By applying the hyperbolic sinusoidal simplification in Section~\ref{sec:ginv_VMF_dist} (labeled "EM-VMF-Hyper"), we can further reduce the computation time by more than a factor of $2$ compared to the original EM-VMF.  

\subsection{$\mathcal G$-invariant Clustering on Simulated Data}
\begin{figure}
  \centering
  \subfigure[VMF Simulated Data]{
    \label{fig:mu_est_2clusters_VMFdata} 
    \includegraphics[width=4.25cm]{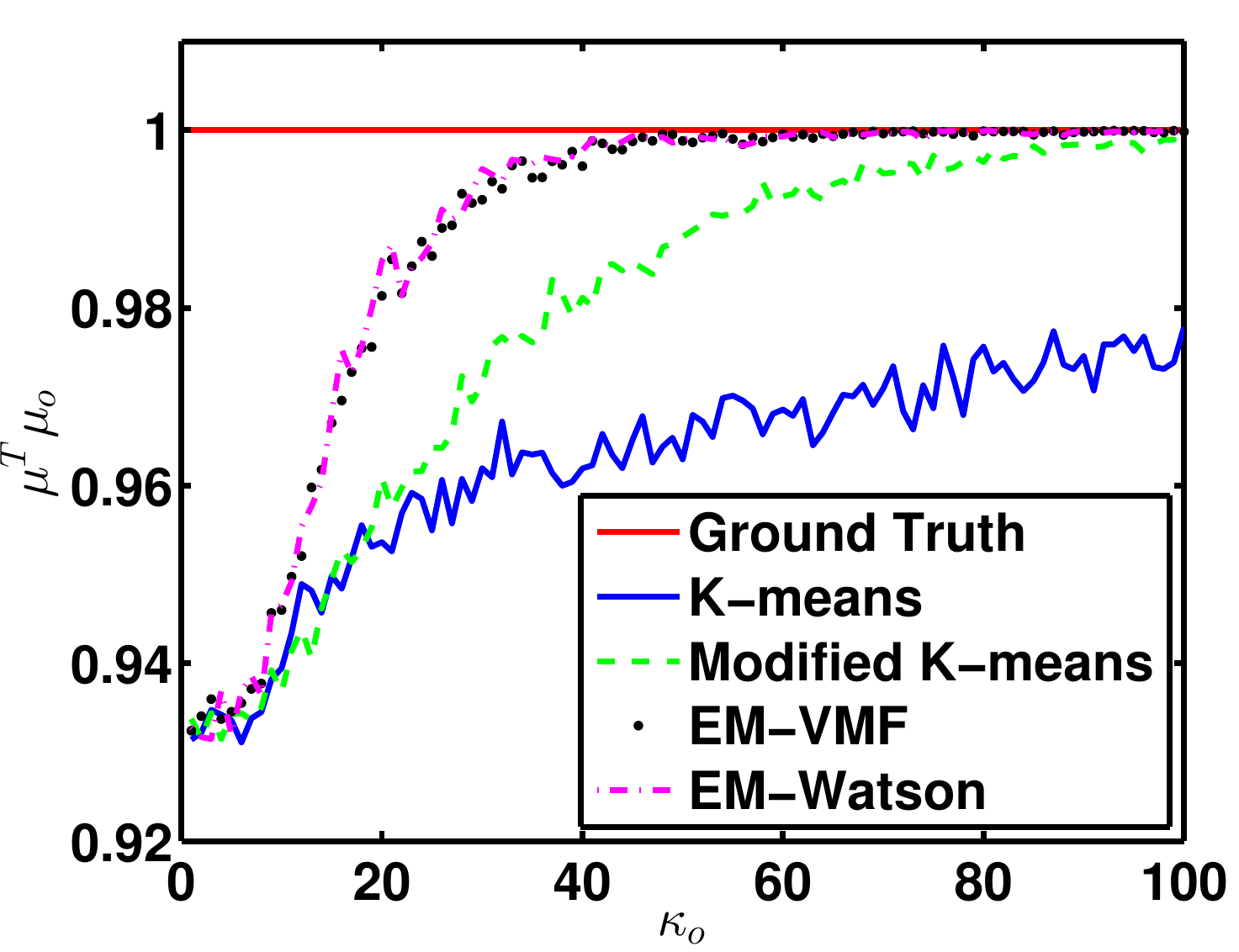}}
  \subfigure[Watson Simulated Data]{
    \label{fig:mu_est_2clusters_Watsondata} 
    \includegraphics[width=4.25cm]{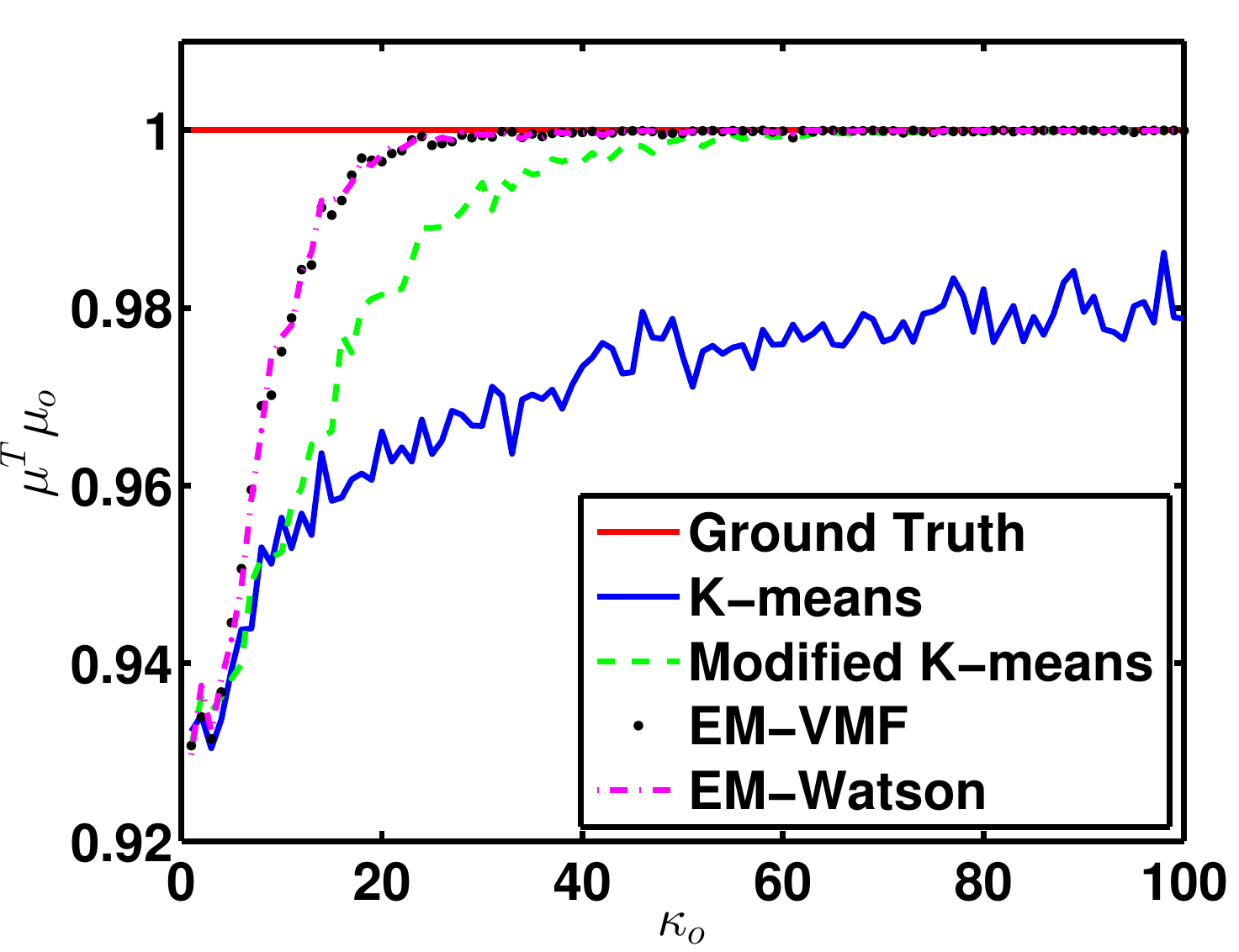}}
  \caption{Mean orientation estimator comparisons for samples generated from two different means. Shown is the average inner product $(\hat{\bmu}_1^T\bmu_1+\hat{\bmu}_2^T\bmu_2)/2$ of four methods when $\bmu_1,\bmu_2$ are the true mean orientations as a function of the true concentration parameter $\kappa_o$ for the data simulated from VMF (Fig.~\ref{fig:mu_est_2clusters_VMFdata}) and from Watson (Fig.~\ref{fig:mu_est_2clusters_Watsondata}) distributions. The K-means with naive estimator ("K-means" in blue line) does not attain perfect estimation for any $\kappa_o$. A modified K-means with ML estimator ("modified K-means" in green dashed line) achieve perfect estimation as $\kappa_o$ becomes large.  The proposed EM-VMF and EM-Watson methods ("EM-VMF", "EM-Watson") achieves perfect estimation much faster than the other methods.}
  \label{fig:mu_est_2clusters} 
\end{figure}

In this section, we demonstrate the performance of our proposed EM approaches for clustering. Sets of $n$ i.i.d. samples were simulated from the VMF or Watson distributions with $\kappa=\kappa_o$ and one of two mean directions ($\bmu_1,\bmu_2$) to generate two clusters of samples. The spherical symmetry group is $m\overline{3}m$ as before. The number of samples for each set was set to $n=1000$ and $\kappa_o$ was swept from $1$ to $100$ while, for each set, $\bmu_1,\bmu_2$ was selected uniformly at random. The experiment was repeated $100$ times and the average values of the inner product $(\hat{\bmu}_1^T\bmu_1+\hat{\bmu}_2^T\bmu_2)/2$ are shown in Fig.~\ref{fig:mu_est_2clusters}. In the figure we compare performances of the following methods: (1) Cluster the samples by standard K-means algorithm with the distance defined by the arc-cosine of the inner product and then use the naive ML within each cluster to estimate the mean directions (labeled "K-means"). (2) 
Cluster the samples by K-means with the distance defined as (\ref{eq:sym_dist}) and then use the aforementioned modified ML estimator (labeled "Modified K-means"). (3) Apply our proposed multi-cluster EM-VMF algorithm to the $n$ samples directly (\ref{eq:mVMF_mClusters_Estep})-(\ref{eq:mVMF_mClusters_Mstep}) (labeled "EM-VMF") (4) Apply our multi-cluster EM-Watson algorithm to the $n$ samples directly (\ref{eq:Watson_mClusters_Mstep_hatT})-(\ref{eq:Watson_mClusters_Mstep_kappa}) (labeled "EM-Watson").

Figure \ref{fig:mu_est_2clusters} shows the average inner product values $(\hat{\bmu}_1^T\bmu_1+\hat{\bmu}_2^T\bmu_2)/2$ from the mean direction estimation. The proposed EM-VMF and EM-Watson are able to correctly cluster the samples and achieve perfect recovery of the two mean orientations much faster than the other K-means approaches. Notice that the region where all the methods fail is larger than the single cluster case since multiple clusters increase the difficulty of parameter estimation. Again, no matter whether the samples are simulated from VMF or Watson distribution, our proposed approaches perform equally well under both cases.

\begin{figure}
  \centering
  \subfigure[VMF Simulated Data]{
    \label{fig:ROC_VMFdata} 
    \includegraphics[width=4cm]{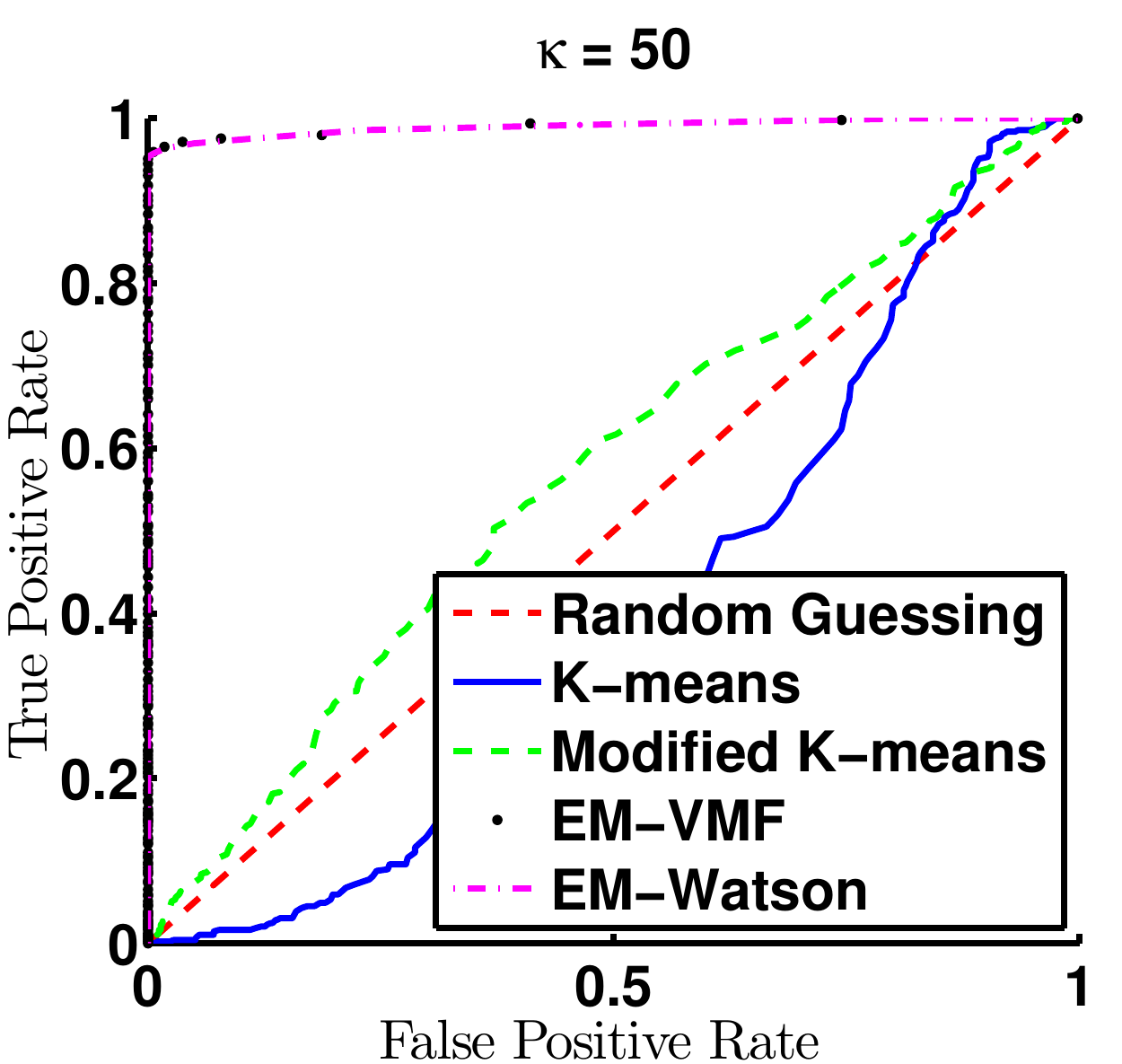}}
  \subfigure[Watson Simulated Data]{
    \label{fig:ROC_Watsondata} 
    \includegraphics[width=4cm]{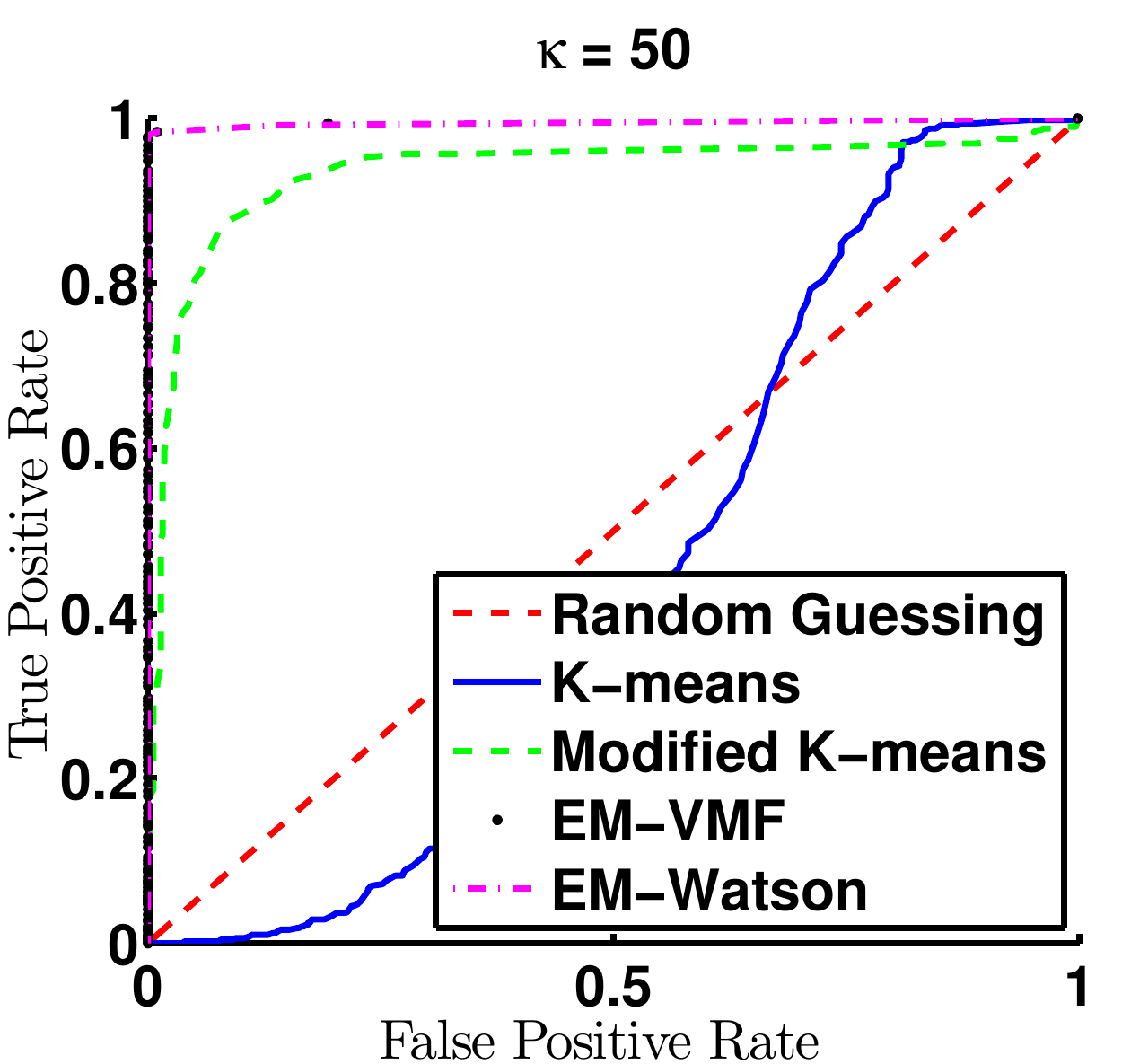}}
  \caption{ROC curve for detecting bi-modally distributed samples. The samples are uni-modal or bi-modal distributed from VMF (Fig.~\ref{fig:ROC_VMFdata}) or Watson (\ref{fig:ROC_Watsondata}) distributions with $\kappa_o=50$. The naive K-means with ML estimator cannot cluster the samples well and estimate the mean directions accurately, resulting in poor detection which is even worse than random guessing. The modified K-means (green dashed line) performs better than K-means but is still unsatisfactory. Our proposed EM-VMF (black dots) and EM-Watson (magenta dashed line) methods have very good performance in this detection task.}
  \label{fig:ROC} 
\end{figure}

To further test the ability to detect multiple clusters given a set of samples, we generate $1000$ sets of samples. Each set has $1000$ samples and is assigned randomly to label $0$ or $1$. If the set is labeled $0$, the samples are generated from a single distribution; If the set is labeled $1$, then the samples in the set are randomly generated from two distributions with different means. The GLRT is used with the four aforementioned clustering methods to test whether the samples in each set are uni-modal or multi-modal. The Receiver Operating Characteristic (ROC) curves of the test results are shown in Fig.~\ref{fig:ROC}. The naive K-means with ML estimator which does not consider the symmetry group actions fails to distinguish whether the multiple modes are from actual multiple distributions or due to the wrap-around effect from the fundamental zone mapping. Therefore, this approach tends to over-estimate the goodness of fit of the $H_1$ model for true negative cases and under-estimate it for true positive cases, resulting in a result that is even worse than random guessing. The modified K-means performs better than K-means but worse than our proposed EM-VMF and EM-Watson algorithms. 

\subsection{EM-ML orientation estimation for IN100 Nickel Sample}
We next illustrate the proposed EM-VMF and EM-Watson orientation estimators on a real IN100 sample acquired from US Air Force Research Laboratory (AFRL) \cite{park_ebsd_2013}. The IN100 sample is a polycrystalline Ni superalloy which has cubic symmetry in the $m\overline{3}m$ point symmetry group.  EBSD orientation measurements were acquired on a $512\times 384$ pixel grid, corresponding to spatial resolution of $297.7$ nm. The Kikuchi diffraction patterns were recorded on a $80\times 60$ photosensitive detector for each of the pixels. 

\begin{figure}[htb]
  \centering
  \subfigure[IPF from OEM]{
  	\includegraphics[width=3.2cm]{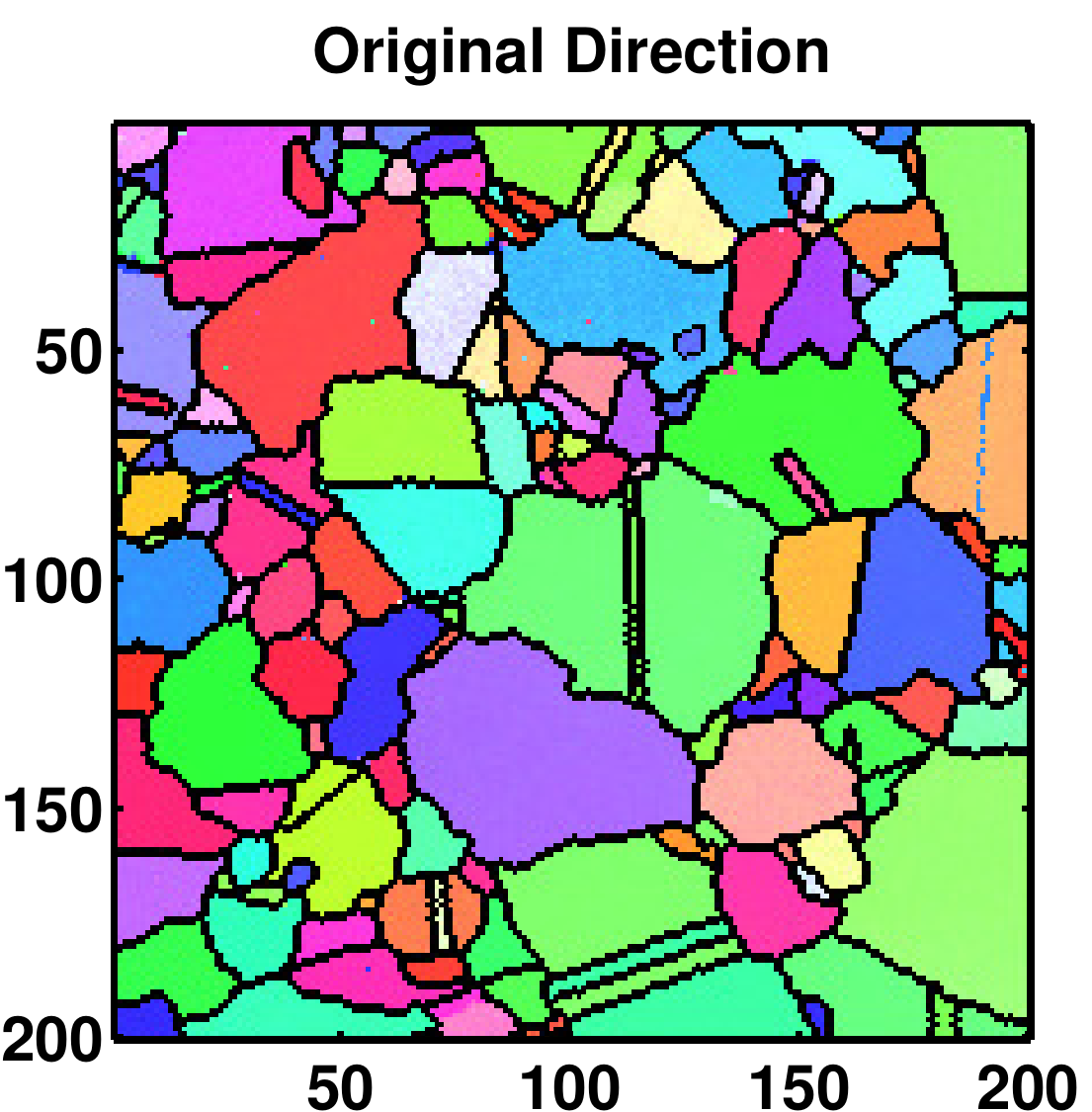}}
  \subfigure[BSE from OEM]{
  	\includegraphics[width=3.8cm]{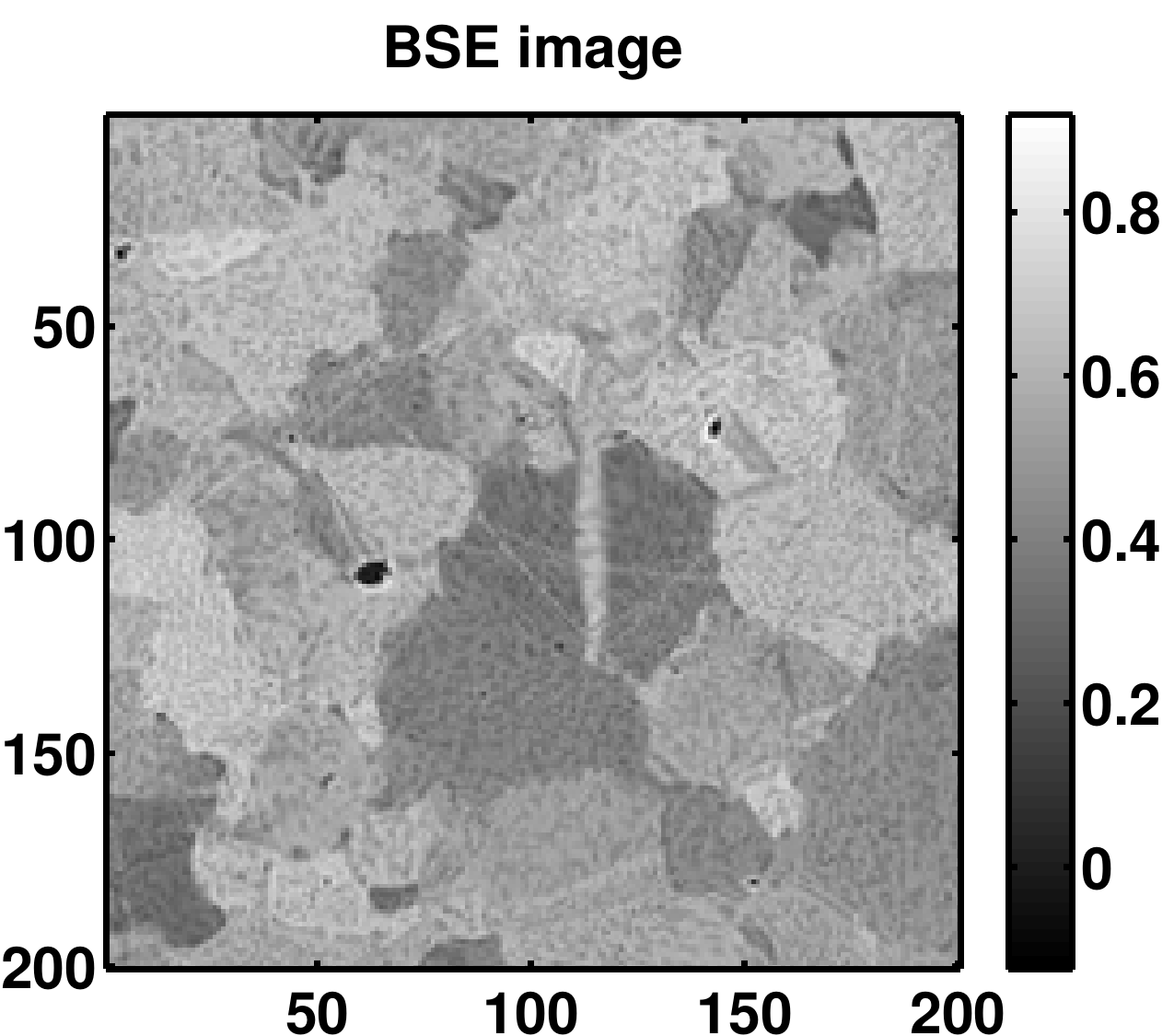}}
  \subfigure[EM-VMF $\hat{\bmu}$]{
  	\includegraphics[width=3.2cm]{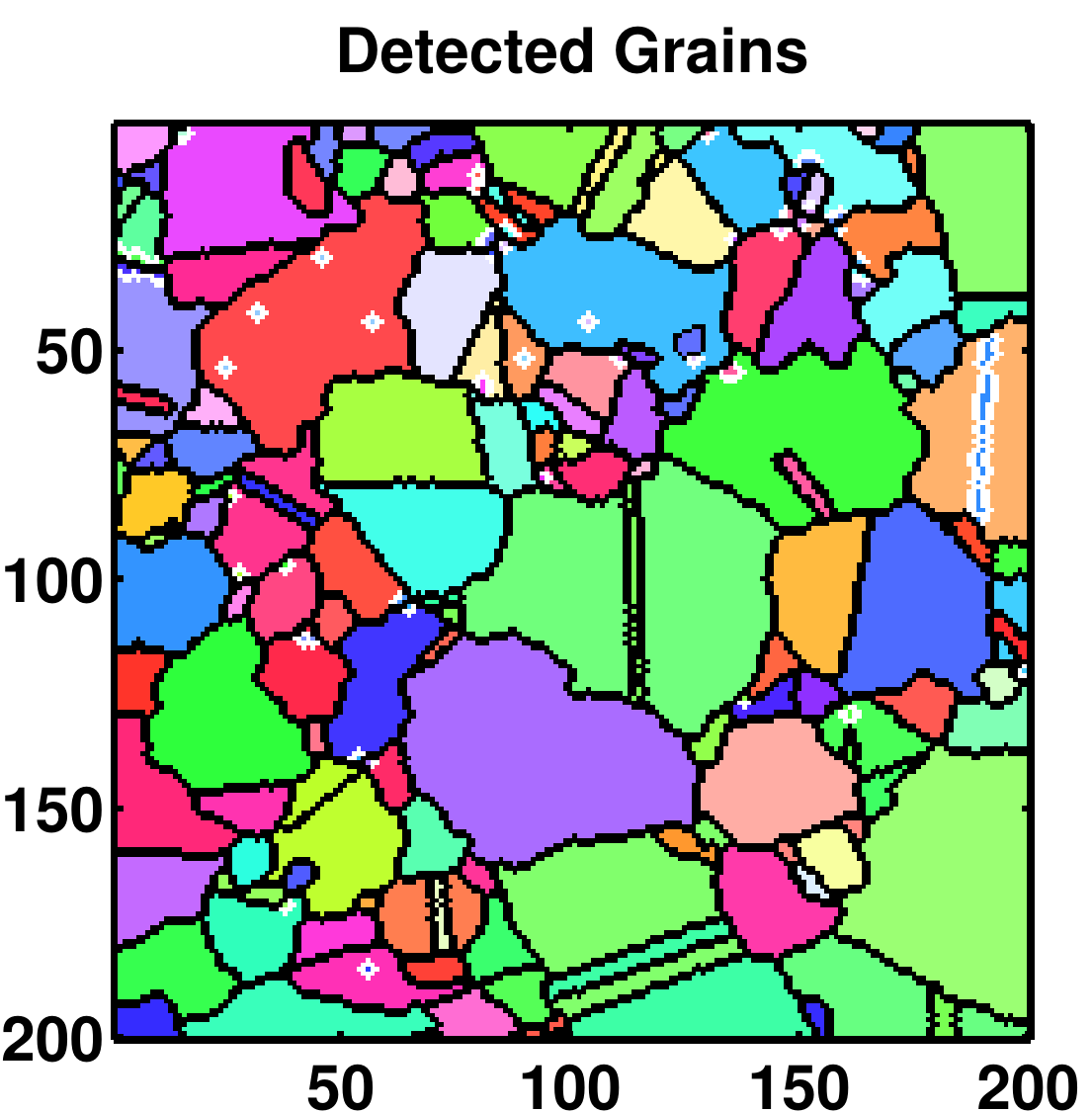}}	
  \subfigure[EM-VMF $\hat{\kappa}$]{
  	\includegraphics[width=3.8cm]{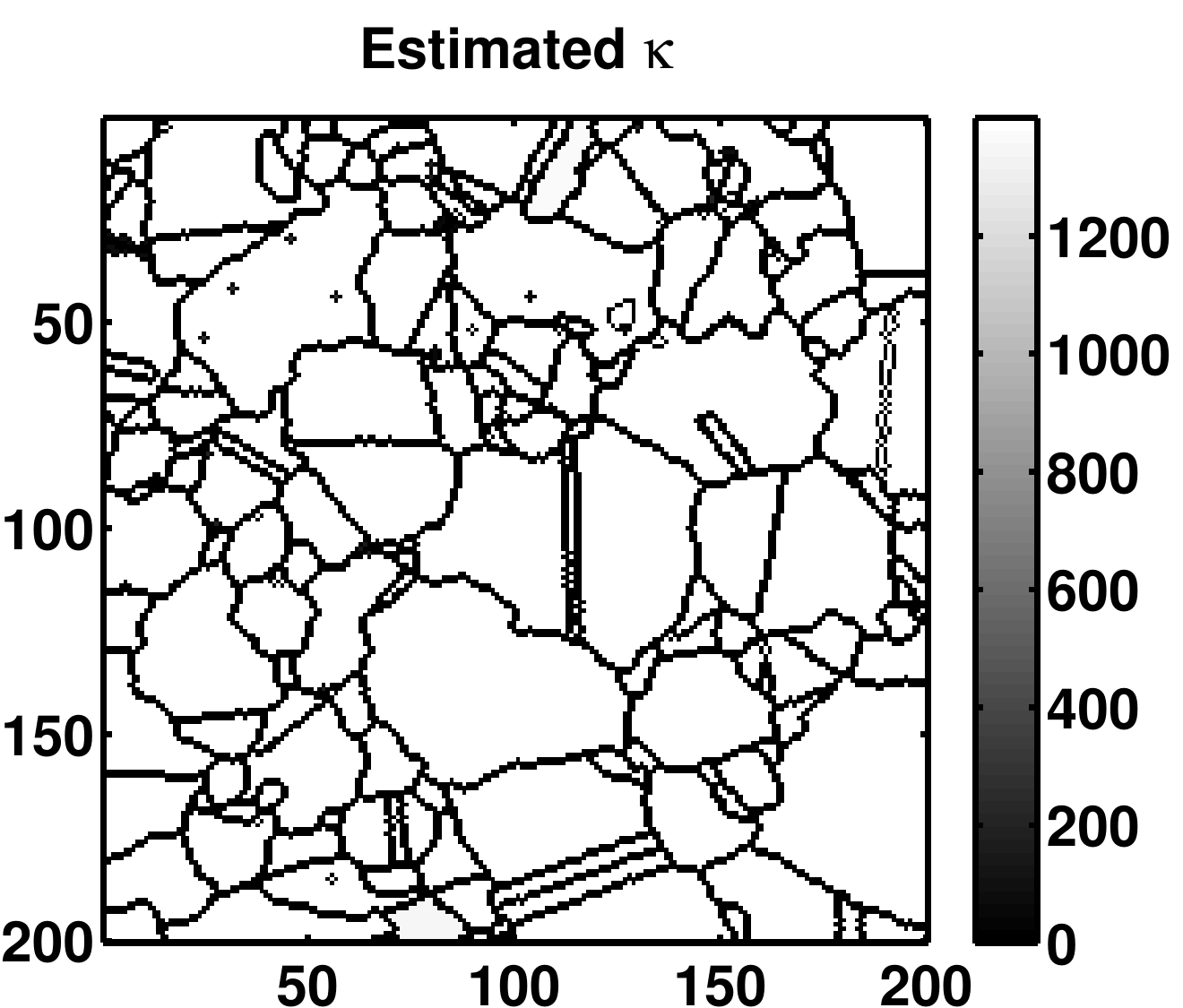}}
  \subfigure[EM-Watson $\hat{\bmu}$]{
  	\includegraphics[width=3.2cm]{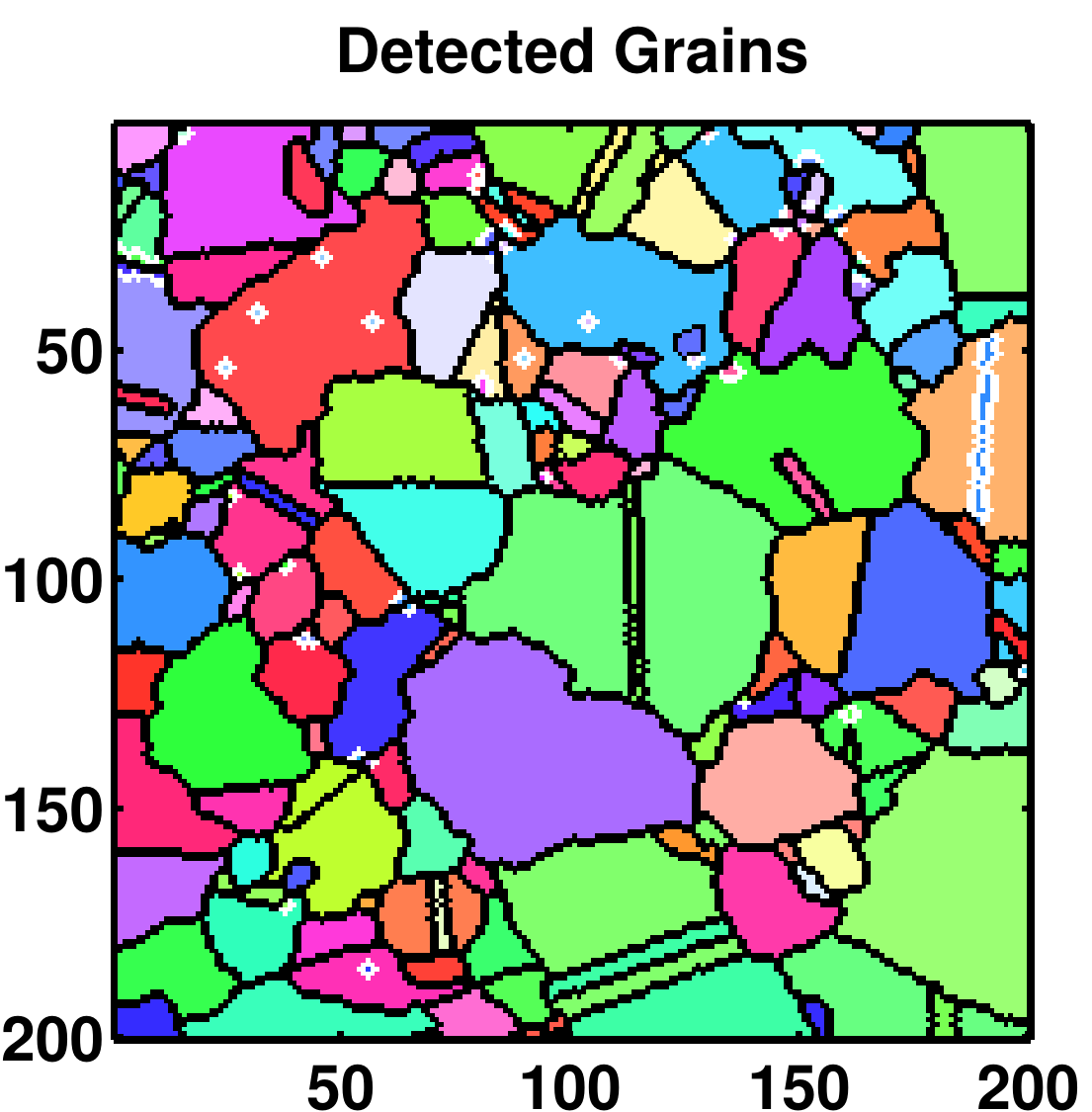}}
  \subfigure[EM-Watson $\hat{\kappa}$]{
  	\includegraphics[width=3.8cm]{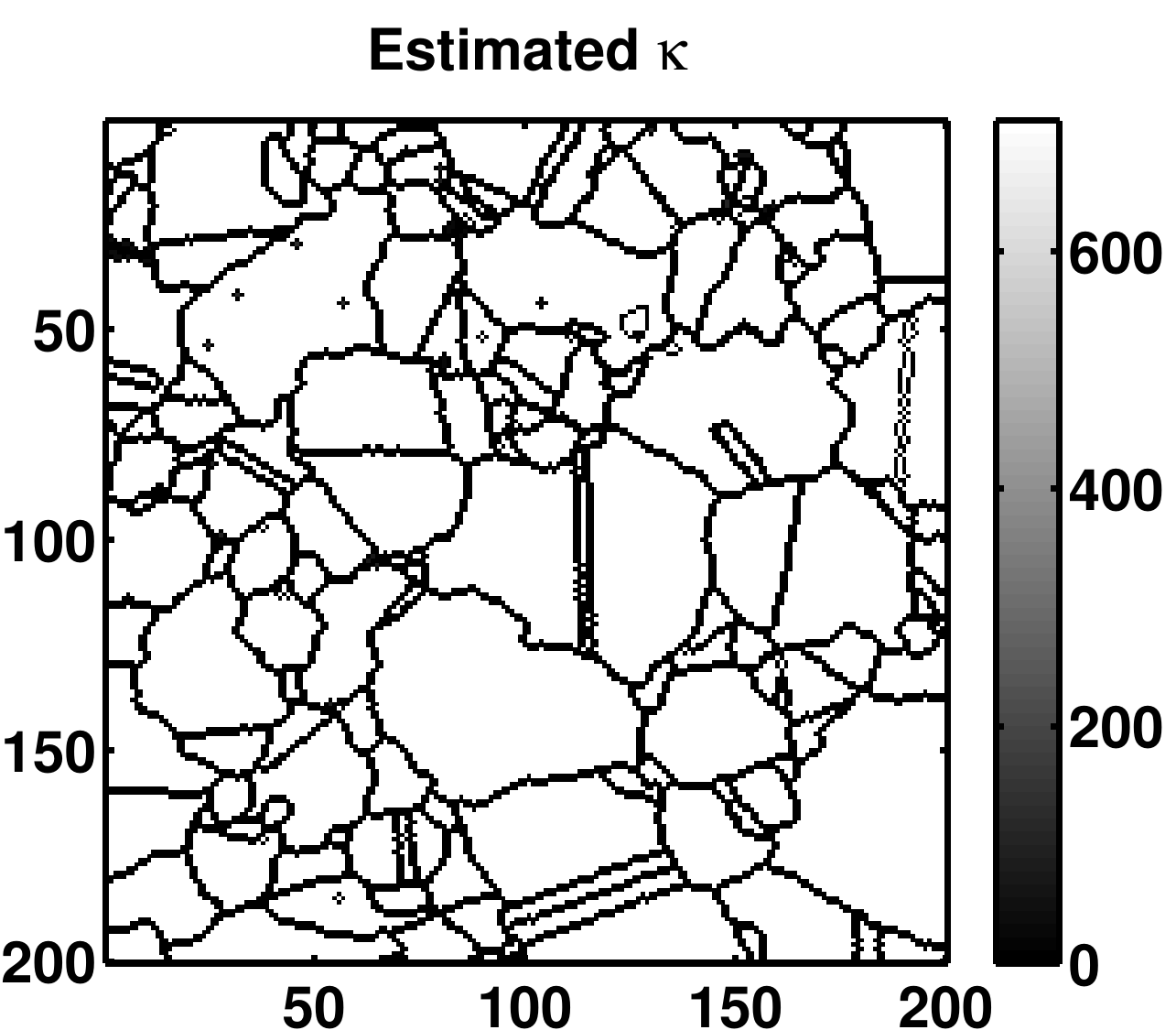}}
  \caption{A $200\times 200$ sub-region of the IN100 sample. (a) is the IPF image for the Euler angles extracted from EBSD by OEM imaging software. IPF coloring in some grains is not homogeneous, likely due to symmetry ambiguity. (b) is the BSE image of the sample. (c)(e) show the estimates of the mean orientations of the regions/sub-regions using a combination of the proposed EM estimators, EM-VMF and EM-Watson respectively, and the GLRT~(\ref{eq:multi_sample_GLRT}) to detect multi-modal distributions within each OEM-segmented region. The sub-regions surrounded by white boundaries indicate those that have been detected as deviating from the distribution of the majority of samples from the same region. (d)(f) show the estimated concentration parameter $\kappa$ for the regions/sub-regions. Note that the estimated $\kappa$ are large for most of the regions/sub-regions because those regions which have multi-modally distributed samples are detected and their concentration parameters are estimated separately for each sub-region.}
\label{fig:IN100}
\end{figure}

Figure \ref{fig:IN100} (a) shows a $200\times 200$ sub-region of the full EBSD sample where the orientations are shown in the inverse pole figure (IPF) coloring obtained from the OEM EBSD imaging software and (b) is the back-scattered electron (BSE) image. Note that the OEM-estimated orientations in some grain regions of the IPF image are very inhomogeneous, having a mottled appearance, which is likely due to a fundamental zone wrap-around problem. As an alternative, we apply a combination of the proposed EM estimators (EM-VMF or EM-Watson) and the GLRT~(\ref{eq:multi_sample_GLRT}) with $C=2$ and significance level $\alpha=0.05$ to detect multi-modal distributions within each OEM-segmented region. Figure \ref{fig:IN100} (c)(e) show the estimates of the mean orientations of the regions/sub-regions, where the sub-regions surrounded by white boundaries indicate those that have been detected as deviating from the distribution of the majority of samples from the same region. The multi-modally distributed regions may be due to undetected grains, inaccurate segmentation, or noisy orientation observations. To distinguish the latter situations from the first in which the region really consists of two grains, the misalignment/noise test introduced in~\cite{chen_coercive_2015} can be used. Figures \ref{fig:IN100} (d)(f) show the estimated concentration parameter $\kappa$ for the regions/sub-regions. Note that the estimated $\kappa$ are large for most of the regions/sub-regions because those regions which have multi-modally distributed samples are detected and their concentration parameters are estimated separately for each sub-region.

\section{Conclusion}
\label{sec:conclusion}
A hyperbolic $\mathcal G$-invariant von Mises-Fisher distribution was shown to be equivalent to the distribution proposed in~\cite{chen_parameter_2015}. The advantage of the hyperbolic form is parameter estimation can be performed with substantially fewer computations. A different group invariant orientation distribution was introduced, called the $\mathcal{G}$-invariant Watson distribution, and an EM algorithm was presented that iteratively estimates its orientation and concentration parameters. We introduced multi-modal generalizations of these $\mathcal G$-invariant distributions using mixture models and showed that these can be used to effectively cluster populations of orientations that have spherical symmetry group invariances. The mixture of VMF and Watson models were applied to the problem of estimation of mean grain orientation parameters in polycrystalline materials whose orientations lie in the $m\overline{3}m$ point symmetry group. Application of the finite mixture representation to other types of groups would be worthwhile future work.

\section*{Acknowledgment}
The authors are grateful for inputs from Megna Shah, Mike Jackson and Mike Groeber. AOH would like to acknowledge financial support from USAF/AFMC grant FA8650-9-D-5037/04 and AFOSR grant FA9550-13-1-0043. MDG would like to acknowledge financial support from AFOSR MURI grant FA9550-12-1-0458.



%

\bibliographystyle{IEEEtran}
\bibliography{Fusion2015}

\end{document}